\documentclass{article}

\usepackage{PRIMEarxiv}

\usepackage[utf8]{inputenc} 
\usepackage[T1]{fontenc}    
\usepackage{hyperref}       
\usepackage{url}            
\usepackage{booktabs}       
\usepackage{amsfonts}       
\usepackage{nicefrac}       
\usepackage{microtype}      
\usepackage{lipsum}
\usepackage{fancyhdr}       
\usepackage{graphicx}       
\graphicspath{{media/}}     
\usepackage[utf8]{inputenc} 
\usepackage[T1]{fontenc}    
\usepackage{hyperref}       
\usepackage{url}            
\usepackage{booktabs}       
\usepackage{amsfonts}       
\usepackage{nicefrac}       
\usepackage{microtype}      
\usepackage{xcolor}         
\usepackage{authblk}
\usepackage[numbers]{natbib}
\usepackage{amsmath}
\usepackage{mathtools}
\usepackage{amssymb}
\usepackage{xparse}
\DeclarePairedDelimiter{\abs}{\lvert}{\rvert}
\DeclarePairedDelimiter{\norm}{\lVert}{\rVert}

\usepackage[inline]{enumitem}   
\usepackage[ruled, lined, linesnumbered, commentsnumbered, longend]{algorithm2e}
\usepackage{algorithmicx}
\usepackage{algpseudocode}
\usepackage{authblk}
\usepackage{comment}
\usepackage{multirow}
\usepackage{graphicx}
\usepackage{bbm}
\title{Error Controlled Feature Selection for Ultrahigh Dimensional and Highly Correlated Feature Space Using Deep Learning
}

\author[1]{Arkaprabha Ganguli\thanks{gangulia@msu.edu}}
\author[2]{David Todem}
\author[1]{Tapabrata Maiti}
\affil[1]{Department of Statistics \& Probability, Michigan State University}
\affil[2]{Department of Epidemiology \& Biostatistics, Michigan State University}

\begin{document}
\maketitle

\begin{abstract}
 In recent years, deep learning has been at the center of analytics due to its impressive empirical success in analyzing complex data objects. Despite this success, most of the existing tools behave like black-box machines, thus the increasing interest in interpretable, reliable, and robust deep learning models applicable to a broad class of applications. Feature-selected deep learning has emerged as a promising tool in this realm. However, the recent developments do not accommodate ultra-high dimensional and highly correlated features, in addition to the high noise level. In this article, we propose a novel screening and cleaning method with the aid of deep learning for a data-adaptive multi-resolutional discovery of highly correlated predictors with a controlled error rate.  Extensive empirical evaluations over a wide range of simulated scenarios and several real datasets demonstrate the effectiveness of the proposed method in achieving high power while keeping  the false discovery rate  at a minimum. 
\end{abstract}


\section{Introduction}
In modern applications (e.g., genetics and imaging studies), the investigator is often interested in uncovering the true pattern between a quantitative response and a large number of features. The key working assumption, oftentimes, is that there is an underlying sparsity pattern buried in the high dimensional data setting.
Selecting the essential features aids in further scientific investigations by offering improved interpretability and explainability, reduced computational cost for prediction and estimation, and less memory usage due to lower dimensional manifolds of the feature space being estimated. 
Under the linear model (LM) framework, this problem has been extensively studied over the past few decades producing popular algorithms such as Lasso, Elastic net, SCAD, and MCP. A detailed review of this literature can be found in  \citet{fan2010selective} and thus is omitted here. However, regardless of their ubiquitous applications, the LM has limited usage, especially when the underlying mechanism is highly nonlinear, with potential interaction effects. 
Relaxing the linearity assumption, the Artificial Neural Network (ANN) models are well known for efficiently approximating complicated functions. From an information-theoretic viewpoint, \cite{elbrachter2021deep} established that deep neural networks (DNN) provide an optimal approximation
of a nonlinear function, covering a wide range of functional classes used in signal processing. This property has promoted the use of Deep Learning (DL) models for feature selection, an approach that has generated much research interest over the past few years. A major caveat, however, is that the DL models are often used as a black box in many applications. Following the intriguing arguments in \cite{rudin2019stop}, caution must be exercised regarding the application of DL models for decision-making in real-world problems. Employing only the relevant predictors to construct a predictive model is obviously a right step toward explainable machine learning. However, as suggested in \cite{fragile_NN}, oftentimes, the feature importance in DL-based algorithms varied drastically under small perturbations in the input or in the presence of added noise. 


As a solution to this problem, we focus on the reproducible nonlinear variable selection using DL models with some error control. We adopt the False Discovery Rate (FDR) first proposed by \cite{benjamini1995controlling}, known for being suitable for large-scale multiple testing problems.
To formally define the FDR, we consider the random variable, $FDP$, representing the False Discovery Proportion: $ FDP= \frac{e_0}{N_{+}} \mathbbm{1}\left(N_{+}>0 \right)$, where $e_0$= number of falsely selected variables and $N_{+}$= number of total discoveries. Then, FDR is defined as $FDR=E(FDP)$. Estimating this expectation poses a unique challenge for the model-free variable selection problem, which many authors have tried to solve from various perspectives. For example, a p-value approach has been proposed as a feature importance criterion in multiple testing literature; see \cite{BB-FDR,NeuralFDRLD,li2019multiple,adapt} for a more detailed overview. However, for DL models, generating interpretable p-values is still an unrevealed research problem. To circumvent this limitation, the knockoff framework has been proposed by \cite{Model_X}. Essentially, this is a model-free variable selection algorithm with provable FDR control, assuming one has prior knowledge of the predictors' distribution.  \cite{deeppink}  further proposed the \textit{DeepPINK} algorithm by integrating the knockoff framework with the DL architecture for improved explainability of the DL models. However, in real-world applications, the predictor's distribution needs to be estimated to generate the knockoff variables, which adds another layer of uncertainty to the analysis. Recently \cite{barber2020robust} showed that the knockoff framework might yield inflation in false discoveries, consistent with the error incurred in estimating the predictor's distribution. This problem is exacerbated for highly correlated features. An empirical illustration is provided in  section 3 of the supplementary material (SM), showing how model-X knockoff \citep{Model_X} typically fails to control FDR under a simplistic setting with high multicollinearity. In some cases, it may be possible to have prior knowledge of the correlation pattern among the features. For example, in genetics studies, there is a common notion of linkage disequilibrium, which helps specify the dependence pattern among the alleles at polymorphisms (\cite{knockoff-hmm}). In other domain sciences, however, this information is typically not available. Hence any model-specific knockoff generation \citep{Model_X, knockoff-hmm} would be inefficient in those contexts. Recently, DL-based flexible knockoff generating algorithms have been proposed \citep{knockoff-autoencoders, knockoffgan,deep-knockoffs}; however they are trained in a typical big-$n$-small-$p$ setting, and it is unclear how they will perform when the sample size $n$ is significantly smaller than the dimension of the covariates $p$, and the predictors are highly correlated. We next discuss in details the multicollinearity issue.

    
In many modern high-dimensional datasets arising in genetics and imaging studies, the other challenge is extreme multicollinearity - the predictors are typically correlated among themselves in a complex manner, often with pairwise sample correlations exceeding 0.99. 
Because extremely correlated features become almost indistinguishable, in a regression setup, it would be unrealistic to claim that only one of the features of a cluster is associated with the outcome. Hence, accounting for the uncertainty, it would be pragmatic to aim for group-level variable selection and claim that at least one variable from a densely correlated group is important for the outcome. 
In this context, the term '\textit{true discovery}' implies that the selected cluster can serve as a good proxy for at least one element in the true index set of significant features. However, a complication of this approach is that the notion of FDR becomes non-trivial. For this reason, following \cite{siegmund2011false}, we adopt the cluster version of the FDR  as the \textit{ expected value of the "proportion of clusters that are falsely declared among all declared clusters"}. We denote this as \textit{cFDR} henceforth. 
Looking at the extreme multicollinearity problem from a slightly different angle, several algorithms have been proposed in the hierarchical testing literature including \textit{CAVIAR} \citep{cavier}, \textit{SUSIE} \citep{susie}, \textit{KnockoffZoom} \citep{knockoffzoom}. While the knockoff-based procedures have the limitation of generating knockoffs from an unknown distribution with a very small sample size, other methods lack applicability in non-linear-nonparametric setups as they typically depend on p-values. 

\paragraph{Our contribution} To address the complications mentioned above in variable selection and unexplored gap while applying DL, we propose SciDNet- Screening \& Cleaning Incorporated Deep Neural Network - a novel method for the reproducible high-dimensional nonlinear-nonparametric feature selection with highly correlated predictors. 
The screening step is a dimension reduction step. We screen out most of the null features and create a set of multi-resolution clusters that collectively contain all the proxy variables needed to cover the truly significant features with high probability. In the cleaning step, using a properly tuned DL model under an appropriate resampling scheme, an estimator of the falsely discovered clusters is proposed, followed by constructing a surrogate of the FDR. Finally, we select some clusters of highly correlated predictors by controlling the surrogate FDR. To this end, 'FDR observed for SciDNet' would implicitly mean the value of the cFDR discussed here. 

To the best of our knowledge, in a high-dimensional setting, no other method in the literature accommodates the multicollinearity issue via data adaptive cluster formation, followed by a nonlinear-nonparametric controlled feature selection integrated with DL. Our thorough empirical study demonstrates the proposed method's validity in general as a proof of concept by achieving higher power, controlled FDR and higher prediction accuracy. The proposed approach doesn't rely on any modelling assumptions and is completely free of p-value unlike existing state-of-the-art methods, providing a better understanding of the sparse relationship between the outcome and the high-dimensional predictors. While theoretically guaranteed FDR control in DL-framework is still an  active area of research, the current study further opens up avenues for a theoretical investigation of a rigorous mathematical foundation for generalizability and broader validity.

For the rest of the article, in Section \ref{sec:algorithm}, we describe the proposed screening and cleaning method, followed by an extensive simulation study in Section \ref{sec:simulation} and analysis of two real-world gene expression dataset in Section \ref{sec:real_data}. Finally, Section \ref{sec:conclusion}  concludes with a summary and future directions.

\section{The Algorithm}\label{sec:algorithm}

\subsection{Notation and assumptions}
Under the supervised learning framework, let $Y$ denote a continuous response variable, and $X = (X_1,\dots, X_p)$ denote p continuous covariates. Let $F_y(\cdot)$ denote the CDF
of the response variable $Y$, and let $F_k(\cdot)$ denote the CDF of the
predictor $X_k$. Assuming a sample size $n$, we consider the ultrahigh-dimensional setting where $p=O(exp\left(n^\tau\right)), \tau>0$. We assume no specific functional relationship between the outcome $Y$ and the predictors $X$ but we impose a high-level assumption on the distribution of X. In the spirit of \cite{liu2009nonparanormal}, we assume that the predictors follow the nonparanomal distribution; i.e., there exist unknown functions $f(X)=\{ f_j(X_j),j \in \{1,2,\dots,p\}\}$, such that $ f(X) \sim N(\mu^{p\times1},\Sigma^{p\times p})$. This nonparanomal distribution covers a wide range of parametric family of distributions and its main beauty lies in the fact that $f(X)$ preserves the conditional dependency structure of the original variables $X$. Maintaining the sparsity condition, we may assume that there exists a subset $S_0 \subset  \{1, 2,\dots , p\}, |S_0|=O(1)$, such that, conditional on features in $S_0$, the response $Y$ is independent of features in $S_0^c$. In other words, $S_0=\{k: f(y|X)$ depends on $X_k\}$, where $f(y|X)$ is the conditional density of y given X. Our goal is to learn the sparsity structure by estimating $S_0$.

\subsection{Screening Step}

Under the assumption that cardinality of $S_0$ is much smaller than the feature space dimension $p$, most of the features belong to $S_0^c$. Hence in the screening step, we focus primarily on finding an active set $\hat{S}_n$ with $|\hat{S}_n|<<p$ such that  $P(S_0 \subset \hat{S}_n ) \to 1$ as $n \to \infty$. This property is called the sure screening property \citep{fan2008sure}, which ensures that all the significant predictors are still retained in $\hat{S}_n$ and the other predictors $\{X_j, j \in \hat{S}_n^c\}$ are henceforth eliminated from the remaining analysis. As these active variables are highly correlated among themselves, in second step we further cluster them by exploiting the conditional dependency structure.

\subsubsection{Finding the active set of variables}
To find the active set, we first consider the nonparanomal transformation on $(Y,X)$ and then perform the Henze–Zirkler’s (HZ) test on the transformed variable. While the first transforms all the variables to a joint  Gaussian variable maintaining their conditional covariance structure; the second test confirms, by pairwise testing, if there are significant dependence in the transformed response and predictors. This workflow has been proposed by \cite{liu2009nonparanormal},\cite{HZ}, \cite{xue2017robust}.  The strategy proceeds as follows:

\begin{enumerate}
    \item  \textbf{Nonparanomal transformation}: We first consider the following transformation: $T_y(Y)= \mathbf{\Phi}^{-1}(F_y(Y)), T_k(X_k)=\mathbf{\Phi}^{-1}(F_k(X_k)), k=1,2,\dots,p$\label{nonparanomal}, where $\mathbf{\Phi}(\cdot) $ denotes the CDF of the standard Gaussian distribution. However, in practice, the cdf of $Y$ and $X_k$ are unknown, we can estimate it by the truncated empirical cdf as suggested by \cite{liu2009nonparanormal}. Henceforth, let $(\tilde{T}_y(Y),\tilde{T}_k(X_k))$ denote the corresponding transformations.

    \item \textbf{HZ test}: By the basic properties of CDF, it is easy to see that $(T_y(Y),T_k(X_k))$ will jointly follow a bivariate Gaussian distribution $N_2(0, I_2)$ if and only if $Y$ is independent of $X_k$. This can be tested using HZ test, \cite{HZ}, where the test statistic for the predictor $X_k$ can be expressed as $w_k=\int_{\mathcal{R}^2} |\psi_k(t)-exp(-\frac{1}{2}t't)|^2\phi_\beta(t)dt, k=1,2,\dots,p$; where $\psi_k(t)$ is the characteristic function of $(T_y(Y),T_k(X_k))$ and $exp(-\frac{1}{2}t't)$ represents the characteristic function of $N_2(0, I_2)$. It typically measures the disparity between the joint distribution of $(T_y(Y),T_k(X_k))$ and $N_2(0, I_2)$ and is expected to be typically high for the non-null predictors $X_j, j \in S_0$ indicating significant evidence against the independence of the transformed variable $(T_y(Y),T_k(X_k))$.  
    
    Next, as in practice, we proceed with $(\tilde{T}_y(Y),\tilde{T}_k(X_k))$, we calculate the  the HZ test statistic as 
    \begin{equation}\label{HZ_statistic}
    \resizebox{0.45\textwidth}{!}{$\tilde{w}_k^*= \frac{1}{n^2}\sum_{i=1}^n \sum_{j=1}^n e^{-\frac{\beta^2}{2}d_{ij}}-\frac{2}{n(1+\beta^2)}\sum_{i=1}^n e^{-\frac{\beta^2}{2(1+\beta^2)}d_i} + \frac{1}{1+2\beta^2}$}
    \end{equation}
    where $d_{ij}=(\tilde{T}_k(x_{ki})-\tilde{T}_k(x_{kj}))^2+(\tilde{T}_Y(y_i)-\tilde{T}_Y(y_j))^2$ and $d_i=\tilde{T}_k^2(x_{ki})+\tilde{T}_Y^2(y_i)$. Consistent with the existing literature, we choose the value of the smoothing parameter $\beta$ as $\frac{(1.25n)^{1/6}}{\sqrt{2}}$, which corresponds to the
    optimal bandwidth for a nonparametric kernel density estimator with Gaussian kernel (\cite{HZ}). The observed test statistics $\tilde{w}_k^*$ converges to $w_k$ as shown in \cite{xue2017robust}.
    \item Next, we select the active set of predictors $\hat{S}_n$ according to the larger values of $\tilde{w}_k^*$, i.e., $\hat{S}_n= \{1 \leq k \leq p : \tilde{w}_k^* > cn^{-\kappa}\} \label{active_set}$
    where where $c$ and $\kappa$ are predetermined threshold values. 
\end{enumerate}
This active set $\hat{S}_n$ contains all the predictors significantly correlated with the response marginally. Under very mild regularity conditions on signal strength of the nonnull predictors where $\min_{k \in S_0} w_k\geq 2cn^{-\kappa}$ with c as a constant and $0\leq \kappa \leq \frac{1}{4}$, the screening process enjoys the advantage of sure screening property, i.e., $P(S_0 \subset \hat{S}_n) \to 1$, as $n \to \infty$. More details on the theoretical guarantee can be found in \cite{xue2017robust}. A common practice is to set the active set size $|\hat{S}_n|$ at $\nu_n=[n/log(n)]$. However, as we further cluster the active variables in the next step, our proposed method is fairly robust in terms of the $|\hat{S}_n|$ as long as we retain most of the significant variables. We propose to select a bigger active set with size proportional to $\nu_n$. 

\subsubsection{Clustering the active predictors using the precision matrix} \label{clustering}

\begin{algorithm}     
    \SetKwInOut{KwIn}{Input}
    \SetKwInOut{KwOut}{Output}

    \KwIn{$\left(X \in \mathcal{R}^{n \times p},Y \in \mathcal{R}^n\right)$, \text{The Active set} $\hat{S}_n$ , $|\hat{S}_n|=p_1<p$}
    \BlankLine
    
    \text{Estimate the precision matrix:} $\hat{\Sigma}^{-1}=(\hat{\sigma}^{ij})_{i,j \in \{1,2,...,p\}}$ \text{ using Nodewise Lasso}
    
    \text{Define the clusters } ${C}_i=\{ j \in  \hat{S}_{n}: \hat{\sigma}^{ij}\neq 0\}, i \in \hat{S}_{n}$
    
    \For{$1 \leq i \leq p_1$}{\For{$1 \leq j \leq p_1, j \neq i$}{ Define $\Omega^{ij}=\left\{corr(X_{C_i},X_{C_j})\right\}= \{\rho(X_l, X_{l'}), l \in C_i, l' \in C_j\} \in \mathcal{R}^{|C_i||C_j|}$
    
        \If{$\max\{\Omega^{ij}\}\geq r $}{$C_i=C_i \cup C_j$
        
        $C_j=\phi$
        
        }
        }
    }
       
    \textrm{Retain only the non-null clusters: } ${C}=\{C_i: C_i \neq \phi, i \in \hat{S}_n\}$
        
   \text{Find the cluster representatives } $\tilde{S}_n=\{R_j, 1 \leq j \leq |{C}|: R_j= \underset{l\in {C}_j}{argmax}\{\tilde{w}_l^{*}\}\}$     
    
    \KwOut{Clusters ${C}_1,{C}_2,...,{C}_{|{C}|}$ and corresponding cluster representatives  $\tilde{S}_n=\{R_j, 1 \leq j \leq |{C}|\}$}
    \caption{Finding clusters and the representatives}
    \label{algo_1}
\end{algorithm} 
As the  Henze–Zirkler test focuses on (pairwise) marginal correlation among the predictors and response, it typically includes the  null predictors with  strong associations with a significant predictor; and thus they are highly correlated among themselves. Hence to reduce the high correlation in the active set, our strategy is to exploit their conditional dependency structure and divide the active variables $\{X_j :j \in \hat{S}_n\}$ into $p_c (<< p)$  non-overlapping clusters: $C_1, C_2,\dots,C_{p_c}$. By sure screening property, with asymptotically high probability, $S_0 \subset \bigcup_{j=1}^{p_c} C_j$. The use of sparse precision matrix to understand the dependence structure in a high-dimensional feature space has been well acknowledged in statistics literature (e.g., \cite{lauritzen1996graphical}, \cite{shojaie2010penalized}) due to its scalability. In some contexts, it brings more insight  compared to the analysis of simple covariance matrix. For example, in human brain, two separate regions can be highly correlated with no direct relation and only due to their strong interaction with a common third region. So, understanding the conditional dependence structure and using it in clustering the brain regions is more informative in the context of understanding the functional connectivity in human brain \citep{precision_human_brain}. Otherwise simple correlation based clustering will result in huge cluster sizes with less interpretable group of brain regions. 

To this end, in order to estimate the precision matrix we implement the nodewise Lasso algorithm \citep{van2014asymptotically} on the transformed variables $(\tilde{T}_y(Y),\tilde{T}_k(X_k)) , k \in \hat{S}_n$. Nodewise Lasso regression is generally entertained to estimate a sparse precision matrix in the context of Gaussian graphical model by performing simultaneous Lasso regression on each predictor. The tuning parameters in each nodewise Lasso are typically selected using cross-validation. More details on this algorithm and its  theoretical guarantees can be found in \citep{high_dim_graph}. Let $\hat{\Sigma}^{-1}$ be the estimated precision matrix by the nodewise Lasso algorithm and $\rho(Z,Z')$ denotes any correlation metric for two random variables $Z$ and $Z'$. Algorithm \ref{algo_1} summarizes the clustering step.

Here not only we are clustering the active predictors, but also select an appropriate representative from each cluster. First, for each active predictor $X_i \in \hat{S}_n$, we collect all the other active predictors conditionally dependent on $X_i$, and make cluster ${C}_i$. Although clustering using the conditional dependence produces smaller clusters, there might be some overlaps owing to the complex association in the original predictor space. Hence, to reduce the excessive intercluster correlation, we  merge all those clusters having maximum correlation greater then some pre-specified threshold r (we typically set r=0.9). Next, each cluster is updated by adding all the other features conditionally dependent with the existing cluster members. Finally to find the appropriate cluster representatives, we focus on the HZ-test statistic $\tilde{w}_k^*$ in \ref{HZ_statistic} which measures the extend of resemblance between the distribution of each nonparanomally transformed variable in a cluster and the null distribution $N(0,I_2)$. So, for cluster $C_i$, we select the variable $R_i= argmax_{j\in {C}_i}\{\tilde{w}_j^*\}$ indicating its strongest association with the response variable compared to the other predictors in the cluster.

\subsection{Cleaning with Deep Neural Network (DNN)} \label{cleaning}

 We start the cleaning step by modeling the response $Y$ and the cluster representatives $X_{\tilde{S}_n}$ obtained through \ref{clustering}. In order to perform the error controlled variable selection, each representative will be assigned an importance score followed by a resampling algorithm to finally control the FDR. 
 
  While it is possible to adopt any other generic sparsity inducing DNN procedure, here we focus on the LassoNet algorithm recently proposed by \cite{LassoNet} for its elegant mathematical frameworks which naturally sets the stage for nonlinear feature selection. To approximate the unknown functional connection, it considers the class of all fully connected feed-forward residual neural
  networks; namely, $ \mathcal{F} = \{f \equiv f_{\theta, W}: x \mapsto \theta^T x + h_W(x)  \}\label{ResNet}$.
Here, $W$ denotes the network parameters, $K$ denotes the size of
the first hidden layer, $W^{(0)} \in \mathcal{R}^{p \times K}$ denotes the first hidden layer parameters, $\theta \in \mathcal{R}^{p}$ denotes the residual layer's weights. In order to minimize the reconstruction error:
the LassoNet objective function can be formulated as:
\begin{equation} \label{LassoNet_objective}
\resizebox{0.5\textwidth}{!}{$\min_{\theta, W} L(\theta, W) +\lambda \norm{\theta}_1\\
\text{subject to } \norm{W_j^{(0)}}_\infty \leq M\abs{\theta_j}, j=1,2,\dots,p$}
\end{equation}
With $L(\theta, W)= \frac{1}{n}\sum_{i=1}^n l(f_{\theta, W}(x_i),y_i)$ as the empirical loss on the training data and $x_i$ as the vector of cluster representatives observed for the $i^{th}$ individual. While the main feature sparsity is induced by the $L_1$ norm on residual layer parameter $\theta$, the second constraint controls the total amount of nonlinearity of the predictors. As mentioned in \cite{LassoNet}, LassoNet can be argued as an extension of the celebrated Lasso algorithm to nonlinear variable selection.

In $L_1$ penalization framework, importance of a specific feature is naturally embedded into the highest penalization level upto which it can survive into the model. So, to measure the importance of each representative, the LassoNet algorithm is executed over a long range of tuning parameter $\lambda_1 \leq \lambda_2 \leq \dots \leq \lambda_r$ on $(Y,X_{\tilde{S}_n})$. In practice, a small value is fixed for $\lambda_1$ where all the variables are present in the model. Then we gradually increase the value of the tuning parameter and stop at $\lambda_r$, where there are no variables present in the model. Next the importance score for the $j$-th cluster is defined as $\hat{\lambda}_j$= maximum value of $\lambda$ up to which the j-th representative exists in the model, and then the following rank statistic is computed: $\mathcal{I}_j= \sum_{j^{'} \neq j} \mathbbm{1}\left(\hat{\lambda}_j \leq \hat{\lambda}_{j^{'}}\right)$ for $j=1,2,\dots,\abs{C}$.
A lower $\mathcal{I}_j$ means that the $j$-th cluster representative stays in the model upto a higher value $\lambda$ implying its high potential as a significant cluster; whereas a higher $\mathcal{I}_j$ indicates the corresponding cluster leaves the model even for a smaller value of $\lambda$ as a consequence of being simply a collection of null features. Hence, we should only focus on the clusters with  lower ranks. Additionally, in order to control the FDR, understanding the behaviour of the predictors under the null distribution is important. In traditional FDR controlling algorithms, this is typically done by generating the p-values. Here, as a p-value free algorithm,  we propose the following resampling based approach:

\begin{enumerate}
    \item Generate B bootstrap versions of the data $\left\{Y^b,X_{\tilde{S}_n}^b\right\}_{b=1}^B$ considering only the cluster representatives $\tilde{S}_n$.
    For each bootstrap version, run the LassoNet algorithm parallelly; and calculate the importance of each representative by measuring $\hat{\lambda}_j^b$= maximum value of $\lambda$ up to which the j-th predictor exists in the model for b-th  bootstrap version, and then the ranks $\mathcal{I}_j^b= \sum_{j^{'} \neq j} \mathbbm{1}\left(\hat{\lambda}_j^b \leq \hat{\lambda}_{j^{'}}^b\right)$\label{imp_scr_bt}.
     
    Therefore, the averaged rank is: $ \Bar{\mathcal{I}}_j=\frac{1}{B}\sum_{b=1}^B  \mathcal{I}_j^b$\label{avg_scr}
   
   \item For an arbitrary threshold $\delta$, we would select the cluster representatives with  averaged rank $\Bar{\mathcal{I}}_j$ lower than $\delta$; so we define, $ N_{+}(\delta)= \sum_{j \in \tilde{S}_n} \mathbbm{1}(\mathcal{\bar{I}}_j \leq\delta)$ representing the number of selected clusters with respect to the cutoff $\delta$.
   
   \item Next, to estimate the expected number of falsely discovered clusters, define $\mathcal{R}^b=\{j:\mathcal{I}_j^b \leq \delta \}$, the number of cluster representatives with higher importance score so that the corresponding rank is lower than the cutoff $\delta$ in the b-th bootstrap version. Additionally, define a neighbourhood $\mathcal{N}(\mathcal{\bar{I}}_j,\kappa)= \{l \in \{1,2,\dots,\abs{C}\} : \abs{\mathcal{\bar{I}}_j-l} \leq \kappa\} \label{neighbourhood} $, for some specific small number $\kappa$.
    \item Further, we estimate the number of falsely discovered clusters and hence an estimator of the FDR can be constructed as $\hat{FDR}(\delta)=\frac{ \hat{e}_0(\delta)}{N_{+}(\delta)}$ where,
    \begin{equation} \label{e0_hat}
              \resizebox{0.45\textwidth}{!}{$ \hat{e}_0(\delta)= \frac{1}{B} \sum_{b=1}^{B} \left\{\sum_{j \in \mathcal{R}^b} \mathbbm{1}(\mathcal{I}_j^b \not\in \mathcal{N}(\mathcal{\bar{I}}_j,\kappa))\right\}$}
    \end{equation}\label{fpr}
    \item The $\hat{FDR}$ is sequentially estimated with $\delta=\mathcal{\bar{I}}_{(1)}, \mathcal{\bar{I}}_{(2)}, \dots, \mathcal{\bar{I}}_{(\abs{C})} $ and the optimum threshold is $\Delta^*= \max\{\delta >0 : \hat{FDR}(\delta)<q\}$\label{opt_threshold}
    for some pre-specific FDR control level $q$. The final selected set of clusters with controlled FDR is given by    $\hat{D}_n= \{C_j, j=1,2,\dots,\abs{C}: \mathcal{\bar{I}}_j \leq \Delta^*\}$\label{final_set}
 \end{enumerate}

The proposed method certainly has a close resemblance with an FDR-controlling approach. The notion of a false discovery is incorporated into the algorithm via the resampling: if a null predictor gets relatively higher importance score, that is most possibly due to that specific bootstrap version which creates the spurious relation, however, that would not be consistent for all the other bootstraps in general. On the other hand, all the bootstrap versions should consistently produce higher importance scores for the significant predictors. As a consequence, the variability in the ranks of the importance scores will be much higher for the null predictors compared to their nonnull counterpart. This notion was introduced in the statistics literature in the last twenty years as bagging methods \citep{ breiman1996bagging, bagging} for reducing variance of a black-box prediction. The proposed method utilizes this phase transition in the feature selection framework to effectively identify the false discoveries; an empirical illustration of which is further provided in Section 2 of the SM.  

\begin{table*}
  \caption{Power and empirical FDR with standard error in parentheses showing the feature selection performance}
  \label{power_FDR_simulation}
  \centering
  \resizebox{\textwidth}{!}{\begin{tabular}{lcl|ccccc|ccccc}
    \toprule
  &  &  & \multicolumn{5}{c|}{$\beta=2$}    & \multicolumn{5}{c}{$\beta=4$}    \\
   \cmidrule(lr){4-13}
    &  &  &\multicolumn{4}{c}{SciDNet}& LassoNet & \multicolumn{4}{c}{SciDNet}& LassoNet \\
    \cmidrule(lr){4-7}\cmidrule(lr){9-12}
    \multicolumn{3}{r|}{FDR control Level $q$} & 0.01 & 0.05 & 0.1 & 0.15 & $-$ & 0.01 & 0.05 & 0.1 & 0.15 & $-$\\
    \midrule
    \multirow{6}{*}{Polynomial} & \multirow{2}{*}{$\sigma^2=1$} & Power & 0.59 (0.17) & 0.97 (0.06) & 0.98 (0.05) & 0.98 (0.05) & 0.33 (0.14) & 0.72 (0.32) & 0.91 (0.12) & 0.97 (0.07) & 0.98 (0.05) & 0.37 (0.17)\\
    & & FDR & 0.01 (0.02) & 0.02 (0.04) & 0.04 (0.06) & 0.10 (0.05) & 0.94 (0.02) & 0.01 (0.02) & 0.01 (0.03) & 0.02 (0.04) & 0.05 (0.06) & 0.95 (0.03)\\
    & \multirow{2}{*}{$\sigma^2=5$} & Power & 0.47 (0.27) & 0.80 (0.20) & 0.86 (0.13) & 0.88 (0.12) & 0.33 (0.10) & 0.56 (0.27) & 0.82 (0.19) & 0.93 (0.09) & 0.95 (0.08) & 0.36 (0.17)\\
     & & FDR & 0.00 (0.00) & 0.04 (0.06) & 0.09 (0.07) & 0.12 (0.08) & 0.95 (0.02) & 0.00 (0.00) & 0.01 (0.04) & 0.04 (0.06) & 0.06 (0.07) & 0.92 (0.04)\\
    & \multirow{2}{*}{$\sigma^2=10$} & Power & 0.28 (0.21) & 0.55 (0.25) & 0.65 (0.20) & 0.71 (0.18) & 0.27 (0.22) & 0.40 (0.35) & 0.86 (0.14) & 0.92 (0.10) & 0.93 (0.08) & 0.29 (0.15) \\
    & & FDR & 0.00 (0.00) & 0.03 (0.05) & 0.11 (0.09) & 0.15 (0.11) & 0.96 (0.03) & 0.00 (0.00) & 0.02 (0.04) & 0.04 (0.05) & 0.07 (0.06) & 0.94 (0.03) \\
    \midrule
    
     \multirow{6}{*}{ReLu} & \multirow{2}{*}{$\sigma^2=1$} & Power & 0.63 (0.36) & 0.90 (0.16) & 0.97 (0.06) & 0.98 (0.05) & 0.75 (0.19) & 0.65 (0.38) & 0.96 (0.07) & 0.97 (0.07) & 0.98 (0.05) & 0.81 (0.16)\\
    & & FDR & 0.00 (0.00) & 0.02 (0.05) & 0.05 (0.06) & 0.09 (0.07) & 0.92 (0.02) & 0.01 (0.03) & 0.05 (0.05) & 0.08 (0.06) & 0.12 (0.06) & 0.90 (0.02)\\
     & \multirow{2}{*}{$\sigma^2=5$} & Power & 0.55 (0.22) & 0.82 (0.14) & 0.89 (0.10) & 0.90 (0.10) & 0.48 (0.26) & 0.61 (0.32) & 0.91 (0.14) & 0.95 (0.10) & 0.96 (0.09) & 0.73 (0.17)\\
    & & FDR & 0.00 (0.02) & 0.04 (0.06) & 0.10 (0.09) & 0.14 (0.09) & 0.95 (0.02) & 0.00 (0.02) & 0.02 (0.04) & 0.06 (0.06) & 0.10 (0.08) & 0.93 (0.01)\\
     & \multirow{2}{*}{$\sigma^2=10$} & Power & 0.37 (0.23) & 0.62 (0.22) & 0.74 (0.15) & 0.79 (0.13) & 0.45 (0.28) & 0.65 (0.32) & 0.90 (0.11) & 0.94 (0.08) & 0.95 (0.08) & 0.68 (0.23) \\
     & & FDR & 0.01 (0.04) & 0.06 (0.08) & 0.08 (0.11) & 0.14 (0.13) & 0.96 (0.02) & 0.01 (0.03) & 0.05 (0.06) & 0.11 (0.06) & 0.14 (0.08) & 0.95 (0.02) \\
      \bottomrule
  \end{tabular}}
\end{table*}

\section{Numerical Illustrations} \label{sec:simulation}

\begin{table*}
  \caption{Test MSE with standard error in parentheses showing the prediction performance before and after the feature selection}
  \label{test_MSE_simulation}
  \centering
  \resizebox{\textwidth}{!}{\begin{tabular}{lc|ccc|ccc}
    \toprule
  &   & \multicolumn{3}{c|}{$\beta=2$}    & \multicolumn{3}{c}{$\beta=4$}    \\
   \cmidrule(lr){3-5}\cmidrule(lr){6-8}
    & &  LassoNet & SciDNet+MLP &  SciDNet+RT &  LassoNet & SciDNet+MLP & SciDNet+RT\\
   \midrule
   \multirow{3}{*}{Polynomial} & $\sigma^2=1$ & 1.041 (0.071) & 1.466 (0.188) & 0.685 (0.077) & 2.974 (0.763) & 1.449 (0.473) & 0.445 (0.027)  \\
    &  $\sigma^2=5$ & 1.731 (0.068) & 1.594 (0.222) & 0.921 (0.088) & 3.093 (0.712) & 1.387 (0.514) & 0.679 (0.092)  \\
    & $\sigma^2=10$ & 3.502 (0.057) & 1.468 (0.342) & 0.952 (0.100) & 3.543 (0.158) & 1.395 (0.455) & 0.837 (0.125) \\
       \midrule
   \multirow{3}{*}{ReLu} & $\sigma^2=1$ & 1.745 (0.105) & 1.284 (0.559) & 0.724 (0.085) & 2.459 (0.115) &2.032 (0.943) & 0.625 (0.092)  \\
    &  $\sigma^2=5$ & 2.239 (0.109) & 1.494 (0.557) & 0.797 (0.091) & 3.216 (0.256) & 1.679 (0.701) & 0.753 (0.119)  \\
    & $\sigma^2=10$ & 3.227 (0.133) & 1.494 (0.668) & 0.875 (0.104) & 5.267 (0.221) & 1.187 (0.635) & 0.742 (0.101)\\

    \bottomrule
  \end{tabular}}
\end{table*}

In this section, the finite-sample performance of SciDNet has been investigated using a wide spectrum of simulation scenarios. We first consider the single index model for the data-generating mechanism, a straightforward yet flexible example of nonlinear models. Here the response is related to a linear combination of the features through an unknown nonlinear, monotonic link function, i.e., $y=g(x'\beta)+\epsilon$. We choose the following two link functions: (1) \textbf{Polynomial}: $g(x)= \frac{x^3}{10}+ 3\frac{x}{10}$ and (2) \textbf{ReLu}: $g(x)= max(0,x)$.

We set $n=400$ and $p=1000$. The coefficients $\beta \in \mathcal{R}^p$ is sparse with the true nonzero locations $S_0=\{50,150,250,350,450\}, s=|S_0|=5$, where $\beta_{S_0^c}=0, \beta_{S_0}\sim N_5(u\beta_0,0.1I^{5\times5})$ with $u=\{\pm 1\}^5$. The value of $\beta_0$ is set at $\beta_0= 2,4$ to incorporate varying signal strength . The random error $\epsilon \sim N(0, \sigma^2)$, with three increasing noise level as $\sigma^2= 1,5,10$. The high dimensional predictors are generated from $X \sim N_p(0, \Sigma)$ 
where the covariance matrix $\Sigma$ is chosen as a Toeplitz matrix with $\Sigma_{ij}=\rho^{|i-j|}$. We set $\rho=0.95$.

Two metrics are calculated to measure the feature selection performance of the algorithms: (1) Power= $\frac{|\hat{D}_n \cap S_0|}{|S_0|}$ and (2) empirical FDR = $\frac{|\hat{D}_n \cap S_0^c|}{|\hat{D}_n|}$. In table \ref{power_FDR_simulation} we compare the performance of SciDNet with the basic prediction-optimal LassoNet algorithm. The results indicate the need for a 'cleaning' while using a DL-based model for feature selection. Our method enjoys quick recovery in power when we gradually increase the error-controlling threshold $q$ from 0.01 to 0.15 while maintaining the number of false discoveries below the required level.

Additionally, in table \ref{test_MSE_simulation}, we present a more data-oriented statistical evaluation for further endorsement of SciDNet's discoveries. From the prediction aspect, one would expect that a prediction model implemented only on a handful of features selected by a successful feature selection algorithm would maintain the similar performance of a model implemented on the whole feature space; in some cases, it might enhance the accuracy. To validate this, we randomly split the whole data into 8:2 for training and testing. First, the SciDNet is implemented in the training part, and then two separate prediction models are used only focusing on the selected features : (1) an MLP - a feed-forward multi-layer perceptron
with two hidden layers and (2) bagged regression tree \citep{breiman1996bagging}. These two experiments are henceforth called as:  \textbf{"SciDNet+MLP"} and \textbf{"SciDNet+RT"}. Next, the test data is used to check the out-of-sample prediction accuracy. The reward for a successful feature selection by SciDNet is indicated by the huge gain in test accuracy observed with a lower Mean Square Error (MSE) on the test data. This experiment demonstrates that a black-box predictive model produces more accurate results when applied on the features selected by SciDNet rather than its implementation on the whole feature space, indicating SciDNet's potential use in both feature selection and prediction.

A further simulation study is also conducted considering several nonlinear models as data-generating processes with gaussian and non-gaussian features. The results are promising and demonstrate that SciDNet maintains a satisfactory power-FDR balance for various complicated nonlinear models with and without interaction terms. The results and more details are presented in section 2 of the SM. The hyperparameter selection and further implementation details of SciDNet are relegated to sections 1 and 4 of the SM, respectively. 

Additionally, to motivate the need for cluster-level feature discovery following a screening step, we implemented two recently proposed DL-based FDR-controlled feature selection methods: DeepPink \citep{deeppink} and SurvNet \citep{survnet}. Our empirical study of these models demonstrates that these algorithms are efficient and competitive, given the availability of huge training data and moderate correlation among the features. However, they typically fail with lower power and excessive false discoveries under the current setting of low sample size with correlated features. This is somehow expected, as these methods are not tailored for handling such huge multicollinearity. For example, with the ReLu setting considered above with $\beta=2, \sigma^2=1, \rho=0.95$, DeepPink and SurvNet resulted in $0.067 (0.125)$ and $0.440 (0.314)$ as power with observed FDR at $0.111 (0.194)$ and $0.742 (0.169)$ respectively, at $q=15\%$ FDR control level. Corresponding standard errors are mentioned in the parenthesis. The detailed result is presented in section 4 of the SM.

\begin{table*}[!t]
      \caption{Drug-sensitive genes identified by SciDNet and related prediction performance }
      \label{CCLE_results}
      \centering
      \resizebox{\textwidth}{!}{\begin{tabular}{|l|c|c|c|c|c|c|c|c|}
      \toprule
       \multirow{2}{*}{Drug} & \multicolumn{2}{|c|}{\# genes (clusters) selected} & \multicolumn{3}{|c|}{Test MSE} & \multicolumn{3}{|c|}{Corr($Y_{Pred}, Y_{Test}$)}\\
     \cmidrule(lr){2-9}
       &   by SciDNet & by LassoNet & LassoNet & SciDNet + MLP & SciDNet + RT &  Lassonet & SciDNet + MLP &SciDNet + RT \\
      \midrule
       \midrule
       Topotecan &  25 (9) & 18469 & 1.25 (0.21) & 1.23 (0.14) & 0.81 (0.16) & 0.47 (0.11) & 0.58 (0.06) & 0.69 (0.07)\\
       \midrule
       17-AAG & 12 (8) & 7152 & 1.04 (0.16) & 1.05 (0.09) & 0.83 (0.15) & 0.20 (0.16) & 0.33 (0.10) & 0.49 (0.10) \\
       \midrule
       Irinotecan & 18 (7) & 17727 & 0.93 (0.20) & 1.09 (0.18) & 0.61 (0.13) & 0.59 (0.10) & 0.63 (0.07) & 0.73 (0.08)\\ 
       \midrule
       Paclitaxel & 18 (8) & 16437 & 1.46 (0.33) & 1.46 (0.23) & 1.11 (0.24)& 0.44 (0.14) & 0.45 (0.11) & 0.59 (0.09)\\
        \midrule
        AEW541 & 12 (10) & 15145 & 0.33 (0.06) & 0.39 (0.09) & 0.27 (0.05)& 0.30 (0.14) & 0.49 (0.10) & 0.47 (0.12)\\
    \bottomrule
    \end{tabular}}
\end{table*}

\section{Real Data Analysis} \label{sec:real_data}
In addition to the simulation studies, we implemented the proposed algorithm SciDNet in the following two publicly available gene-expression data sets. We substantiate the findings in two ways: We first provide supporting evidence from the domain research. Additionally, as a more data-aligned validation, we demonstrate that several generic prediction models significantly gain in test accuracy when applied only on the few features selected by SciDNet compared to the prediction result considering the whole feature space. For this purpose, consistent with the other genomic studies, we use the prediction correlation \textit{Corr($Y_{Pred}, Y_{Test}$)} in addition to the test MSE as a metric to measure the test performance. To overcome the extra burden of the low sample size and ultrahigh dimensionality in these data sets, we consider 50 independent replications where the data is divided into training and testing maintaining $8:2$ ratio to get the metrics for the test performance.  The final estimate is obtained by averaging all the test MSEs calculated on each of these replications. Similar approach is considered for the correlation metric as well.

\subsection{Selection of Drug Sensitive Genes using CCLE dataset} 

\begin{figure} 
  \centering
  \begin{minipage}[b]{0.45\textwidth}
    \includegraphics[width=\textwidth]{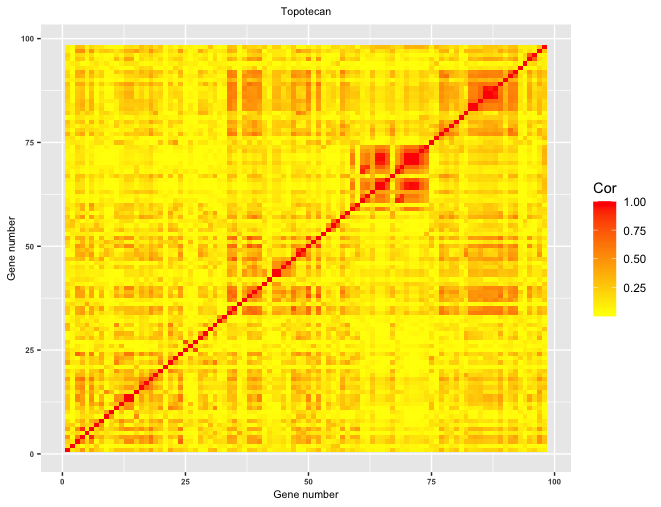}
  \end{minipage}
  \hfill
  \begin{minipage}[b]{0.45\textwidth}
    \includegraphics[width=\textwidth]{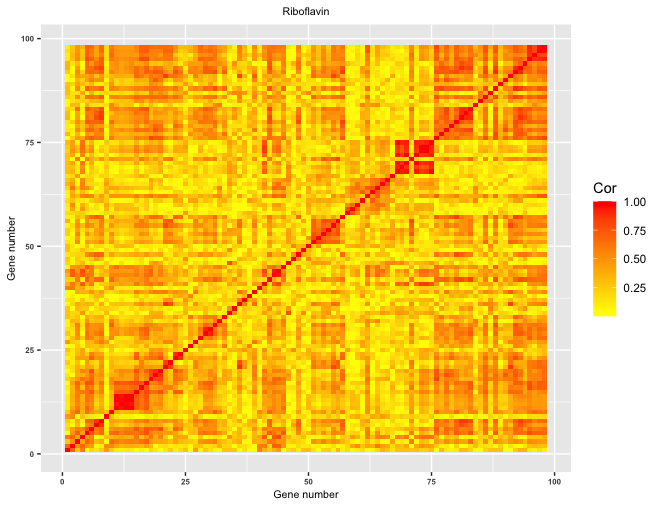}
  \end{minipage}
  \caption{A snapshot of correlation strength for first 100 genes considered for the drug Topotecan in CCLE dataset (left) and riboflavin dataset (right)}
  \vspace{-.2cm}
  \label{heatmap_gene}
\end{figure}

A recent large-scale pharmacogenomics study, namely, cancer cell line encyclopedia
(CCLE, \href{www.broadinstitute.org/ccle}{link available here}), investigated multiple anticancer drugs over hundreds of cell lines. Its main objective is to untangle the response mechanism of anticancer drugs which is critical to precision medicine. The data set consists of dose-response curves for 24 different drugs across over $n=400$ cell lines. For each cell line, it consists of the expression data of $p=18,926$ genes, which we consider as features. For the response, we used the activity area \citep{barretina2012cancer} to measure the sensitivity of a drug for each cell line. Here we seek to uncover the set of genes associated with the following five specific anticancer drugs sensitivity: Topotecan, 17-AAG, Irinotecan, Paclitaxel and AEW541. These drugs have been used to treat ovarian cancer, lung cancer and other cancer types. Previous research outputs on these drugs and related gene expression data can be found elsewhere \citep{barretina2012cancer}.

SciDNet produces multi-resolutional clusters of genes for each of the five drugs considered which are interpretable from the domain science perspective. For example, SciDNet discovers \textit{SLFN11} as the top drug-sensitive gene for the drugs Topotecan and Irinotecan. This is consistent with the previous findings as \cite{barretina2012cancer, zoppoli2012putative} reported the gene \textit{SLFN11} to be highly predictive for both the drugs. For another drug 17-AAG, SciDNet discovers the gene \textit{NQO1} as the topmost important gene which is known to be highly sensitive to 17-AAG \citep{hadley2014use}. The full table containing all the genes selected by SciDNet at $q=15\%$ error-control level and relevant findings from previous genomic studies have been relegated to the section 5 of the SM. Furthermore, similar to the simulation study in section \ref{sec:simulation}, we separately implement the LassoNet on the training data for its simultaneous sparsity-induced prediction-optimal characteristics. The summary of the results are presented in table \ref{CCLE_results} which indicates several interesting points. First, in order to get the prediction optimal result, LassoNet fails to capture the sparsity and discovers a huge number of genes. This is somehow expected as most of the prediction-optimal sparse methods tend to select a larger set of features to maintain the prediction quality \citep{wasserman2009high}. On the other hand, SciDNet produces only $\sim 10$ clusters with an average cluster size $\sim 2.5$. Even with this huge dimension reduction, the added gain in test MSE and \textit{Corr($Y_{Pred}, Y_{Test}$)} further proves that ScidNet successfully retains all the significant predictors. One would further notice, SciDNet+RT achieves the best stable performance which is consistent for all the drugs and our simulation study as well.

\subsection{Selection of associated genes in Riboflavin production data set}
\begin{table}
  \caption{Prediction performance of SciDNet for  Riboflavin production data set}
  \label{riboflavin_test_MSE}
  \centering
   \begin{tabular}{|l|c|c|}
    \toprule
    Algorithm    & Test MSE  & Corr($Y_{Pred}, Y_{Test}$)  \\
    \midrule
    LassoNet	& 0.83 (0.18) & 0.36 (0.30) \\
    SciDNet + LassoNet & 0.19 (0.15) & 0.89 (0.12) \\
   SciDNet + RT & 0.42 (0.16) & 0.74 (0.15)     \\
    SciDNet + MLP     & 2.64 (0.79) & 0.28 (0.37)     \\
    \bottomrule
  \end{tabular}
\end{table}

We further implement the SciDNet in the context of riboflavin (vitamin B2)  production with bacillus subtilis data, a publicly accessible dataset available in the `hdi' package in R. Here the continuous response is the logarithm of the riboflavin production rate, observed for $n=71$ samples along with the logarithm of the expression level of $p = 4088$ genes which are treated here as the predictors. Unlike the previous CCLE data, significant multicollinearity is present in most part of the Riboflavin data set, as demonstrated in Figure \ref{heatmap_gene}. Hence, in order to determine  which genes are important for the riboflavin production, SciDNet resulted in finding 9 clusters of total 160 correlated genes at the $q=15\%$ FDR control level, making the average cluster size of $\sim 17.78$, which is much bigger compared to the previous analysis.  SciDNet discovered the gene \textit{YCIC\_at} as one of the expressive genes related to riboflavin production which was identified by \cite{buhlmann2014high} as a causal gene in this context. The full list of selected cluster of genes by SciDNet are relegated to the section 5 of the SM.

The results from the empirical evaluation of SciDNet's feature selection is presented in table \ref{riboflavin_test_MSE}.  However, for the empirical evaluation, SciDNet+MLP is performing poorly as the inflated cluster dimensions make the input layer of the MLP comparatively large where the number of training data point is $\approx 57$. This necessitates the need for a sparse model here, and we adopt the idea of \textit{Relaxed Lasso}, first proposed by \cite{Meinshausen2007RelaxedL}. Here we implement the LassoNet again on the selected features by SciDNet, which certainly improves the prediction accuracy. Consistent to the previous experiments, SciDNet+RT effectively maintains its prediction performance. This further consolidates the need for applying an apt feature selection method prior to fit a predictive model for an explainable research outcome.

\section{Discussion} \label{sec:conclusion}


 
 While the explainable AI is the need of the hour, statistical models coupled with cutting-edge ML techniques have to push forward because of their solid theoretical foundation clipped with principled algorithmic advancement. The proposed method SciDNet efficiently exploits several existing tools in statistics and ML literature to circumvent some of the complexities that current DL-based models fail to address properly. The basic intuition and exciting empirical results of SciDNet on simulated and real datasets open avenues  for further research. For example, one may be interested in developing a theoretical foundation for this 'screening' and 'cleaning' strategy for provable FDR control. It would be worth mentioning that although we used the sure independence screening with HZ-test and LassoNet as the main tools, SciDNet puts forward a more generic framework and can be implemented with any other model-free feature screening method and sparsity-inducing DL algorithms like \cite{feng2017sparse}. In the screening part, a further methodological extension would consider relaxing the assumption of nonparanomally distributed features for a more flexible approach.  Additionally, after the screening step,  as the dimensionality is reduced, it would be interesting to implement a model-free knockoff generating algorithms like \cite{deep-knockoffs} in the cleaning step as further algorithmic development.  One limitation is that we mainly focus on the regression setup with the continuous outcome because of the requirements of HZ sure Independence test used in the screening step. For a classification task, any model-free feature screening method like \cite{zhou2018model} can be applied in a more general framework.


\bibliography{references_aistat}
\newpage
\appendix

\textbf{Supplementary Material: Feature Selection integrated Deep Learning for Ultrahigh Dimensional and Highly Correlated Feature Space}
\section{Additional Technical Details}\label{details}

In this section, additional technical and implementation details on the proposed algorithm SciDNet are provided. We first show that in the context of multiple testing, under simplistic assumption, $E(\frac{E(e_0)}{N_+}) \geq E(\frac{e_0}{N_+})=FDR$, given that $N_+>0$, where $e_0$= number of falsely selected variables and $N_+$= number of total discoveries. Hence, we provide an upper bound $E\left(\frac{E(e_0)}{N_+}\right)$, which is more tractable than the FDR. We further show our proposed error bound is an estimate of this upper bound of the FDR. Explanation of these notations can be found in Section 2 of the main manuscript.

\subsection{Estimating an upper bound of the FDR}

Let $e_0$ be the number of $\text{false discoveries}; e_0 = \#\{j \not\in S : \mathcal{\bar{I}}_j \geq \delta \}$ and $N_+ =\text{\# total discoveries } =\#\{j \ni \mathcal{\bar{I}}_j \geq \delta \}$,  for a generic cutoff $\delta$. 
Let T denote the number of true discoveries, i.e., $T=\#\{j \in S : \mathcal{\bar{I}}_j \geq \delta \}$ and 
\begin{equation}\label{20}
    \begin{aligned}
        FDR=E(\frac{e_0}{N_+})=E(\frac{e_0}{e_0+T})=E[E(\frac{e_0}{e_0+T}\mid T)] \leq E[\frac{E(e_0\mid T)}{E(e_0\mid T)+T}]
    \end{aligned}
\end{equation}
The last inequality can be obtained by applying conditional Jensen's inequality to the concave function $f(x)=\frac{x}{x+t}$, for  $t>0$.
Now, to show $\frac{E(e_0)}{N_+}$ as an upper bound of the FDP, we proceed as follows:
\begin{equation}\label{21}
    \begin{aligned}
        E[\frac{E(e_0)}{N_+}]=E(e_0)E(\frac{1}{e_0+T})=E(e_0)E[E(\frac{1}{e_0+T}\mid T)]\geq E(e_0) E(\frac{1}{E(e_0 \mid T)+T})\\
        =E(E(e_0\mid T))E(\frac{1}{E(e_0 \mid T)+T})\\
        \geq E(\frac{E(e_0\mid T)}{E(e_0 \mid T)+T})
        \geq FDR
    \end{aligned}
\end{equation}
While the first inequality in \eqref{21} can be obtained by applying conditional Jensen's inequality again to the convex function $f(x)=\frac{1}{x}$ and the third inequality is from \eqref{20}; the second inequality can be justified under the assumption $cov(E(e_0 \mid T), \frac{1}{E(e_0 \mid T)+T} ) \leq 0 \Rightarrow cov(E(e_0 \mid T), \frac{1}{E(N_+ \mid T)} ) \leq 0$. This assumption is fairly intuitive too, stating that conditional on the number of true discoveries, the number of false discoveries and inverse of the number of total discoveries should be negatively correlated.

Hence, we further focus on estimating $E(e_0)$ by $\hat{e_0}$ followed by final FDR-like error bound $\frac{\hat{e_0}}{N_+}$ as an estimate of $E(\frac{E(e_0)}{N_+})$.

\paragraph{Estimating the number of false discoveries}
 At the screening step, the clusters $C_1,C_2,\dots,C_{\abs{C}}$ are formed in such a way that the intra-cluster correlations are not high and thus the cluster representatives are weakly correlated among themselves. So, analogous to p-value, for unimportant clusters, the importance scores will closely follow an uniform distribution. Assuming that $\abs{S_0}=s$,

\begin{equation}\label{E(e_0))}
    \begin{split}
     e_0(\delta)= & \text{Expected number of false discoveries wrt cutoff }\delta\\
     & =E\left(\sum_{j=1}^{\abs{C}} \mathbbm{1}\left(\mathcal{\bar{I}}_j \leq \delta,j \in S_0^c \cap \tilde{S}_n\right)\right)\\
     & \leq E\left(\sum_{j=1}^{\abs{C}} \mathbbm{1}\left(\mathcal{\bar{I}}_j \leq \delta,\mathcal{I}_j^b \sim U\left(s+1,\abs{C}\right)\right)\right) \text{ for any } b=1,2,\dots,B\\
     & = E\left(\sum_{j=1}^{\abs{C}} \mathbbm{1}\left(\mathcal{\bar{I}}_j \leq \delta,\mathcal{I}_j^b \sim U\left(s+1,\abs{C}\right),  \mathcal{I}_j^b \in \mathcal{N}(\mathcal{\bar{I}}_j,\kappa)\right) \right) + \\ 
     & \hspace{2cm}E\left(\sum_{j=1}^{\abs{C}} \mathbbm{1}\left(\mathcal{\bar{I}}_j \leq \delta,\mathcal{I}_j^b \sim U\left(s+1,\abs{C}\right), \mathcal{I}_j^b \not\in \mathcal{N}(\mathcal{\bar{I}}_j,\kappa)\right) \right)\\
     & \leq  2*E\left(\sum_{j=1}^{\abs{C}} \mathbbm{1}\left(\mathcal{\bar{I}}_j \leq \delta,\mathcal{I}_j^b \sim U\left(s+1,\abs{C}\right), \mathcal{I}_j^b \not\in \mathcal{N}(\mathcal{\bar{I}}_j,\kappa)\right) \right)\\
     & \overset{\wedge}{=}\frac{2}{B}\sum_{b=1}^B\left\{\sum_{j \in \mathcal{R}^b} \mathbbm{1}(\mathcal{I}_j^b \not\in \mathcal{N}(\mathcal{I}_j,k))\right\} = \hat{e}_0(\delta)\\
\end{split}
\end{equation}

Hence, for a fixed cutoff $\delta$, our proposed error bound can be considered as an surrogate estimate of an upper bound of the FDR. This is further supported by our empirical study on simulated and real datasets, that SciDNet successfully retains all the important variables while minimising the number of false discoveries and maintains the prediction accuracy even after tremendous dimension reduction. 

\subsection{Hyperparameter Selection}\label{hyperparameters}
   \begin{figure}
      \centering
     \includegraphics[height=8cm,width=14cm]{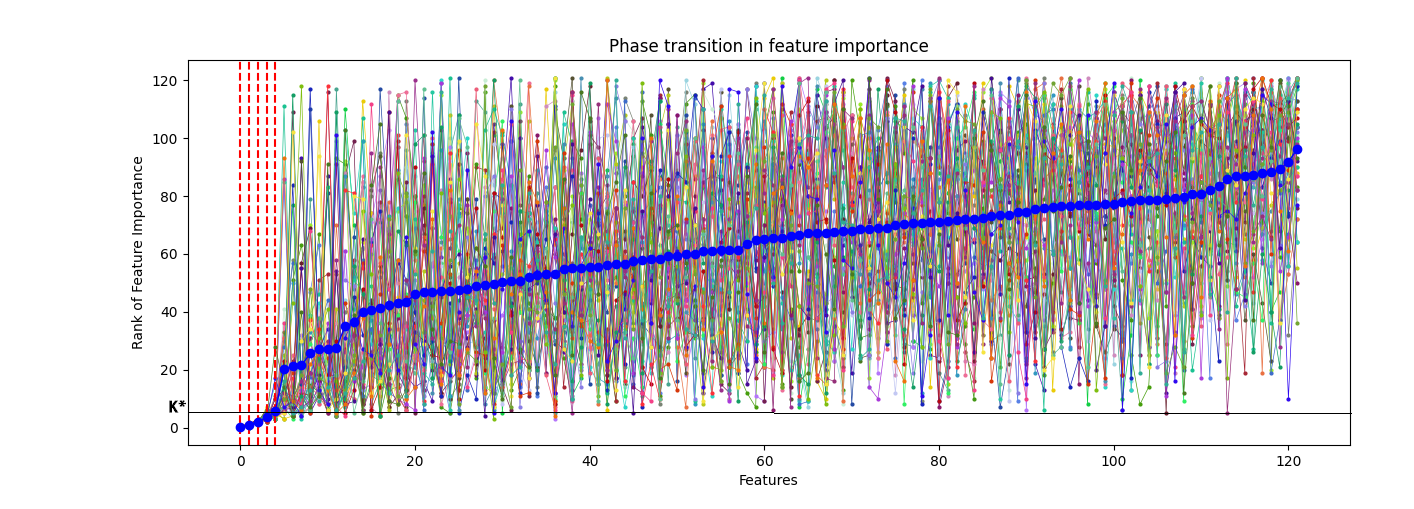}
      \caption{Illustration of phase transition in ranks of feature importance scores}
      \label{fig:fig2}
     \end{figure}
\vspace{-.3cm}
Recently developed Deep Learning (DL) models are generally governed by several hyperparameters and properly tuning them is necessary to get effective results. The proposed SciDNet relies on the following hyperparameters: (1) size of the active set $\hat{S}_n$, (2) the intracluster correlation bound $r$, (3) LassoNet tuning parameters $\lambda$ and $M$ and (4) $\kappa$ used in neighbourhood selection in cleaning step. SciDNet is fairly robust to most of the associated hyperparameters. We discuss a practical way to tune all these hyperparameters here:
\begin{enumerate}
    \item To choose the size of the active set,  we propose to select a bigger active set with size proportional to $\nu_n=[n/log(n)]$. As we further cluster the active variables in the clustering step, a slightly bigger active set with boost up the confidence of sure screening property, See the section \ref{implementation} for an example.
    
    \item  After clustering, the intra-cluster correlation bound $r$ should be fixed at some higher value (usually at $0.9$ or $0.95$) otherwise the cluster sizes will be inflated. 
    \item In cleaning step, a thorough grid search has been done over $\lambda$ considering  $\lambda_1 \leq \lambda_2 \leq \dots \leq \lambda_r$;  in practice, a small value is fixed for $\lambda_1$ where all the variables are present in the model. Then the value of the tuning parameter gradually increased up to $\lambda_r$, where there are no variables present in the model. The other hyperparameter for LassoNet is the hierarchy coefficient $M$ for which we follow the path considered in \cite{LassoNet} and set $M=10$. However, a more flexible approach would be a parallel grid search for $M$ as well. 
    
    \item Finally, choosing an appropriate value of $\kappa$ has a significant effect on the performance of SciDNet. The higher value of $\kappa$ might lead to weaker control over the inclusion of false discoveries, whereas choosing a small $\kappa$ will stricten the error control resulting in reduced power. However, we propose an effective way to tune the $\kappa$ with the assist of phase transition in the ranks of the importance score $\mathcal{\bar{I}}_j$ of the cluster representatives. For an illustration, in Figure \ref{fig:fig2} a simple
    nonlinear additive model (see section \ref{more_simulation_gauss} for more details of the model) is considered with 122 active representatives obtained from the screening step. The first 5 representative features are the only relevant predictors (indicated by the vertical dotted red line) and the ranks of their importance scores are presented along the y-axis. Here the dense lines are the bootstrap ranks $\mathcal{I}_j^b$, observed at 50 bootstrap replications and the solid blue line represents their averaged rank $\mathcal{\bar{I}}_j$ for all the 122 representatives. One would observe a clear phase transition in the bootstrap distribution of the ranks. For the significant features, the ranks are lower with extremely precise estimates whereas for the rest of the null-features, the averaged ranks posses much higher values coupled with huge variability. Hence, for a compact neighbourhood  $\mathcal{N}(\mathcal{\bar{I}}_j, \kappa)$ to capture only the small variability in the bootstrap ranks of the significant features, we simply fix $\kappa$= \textbf{K*} (in figure \ref{fig:fig2}), the phase transition point for the averaged rank.  
 \end{enumerate}

\subsection{Phase transition observed for the CCLE data}     
\begin{figure} 
  \centering
  \begin{minipage}[b]{0.45\textwidth}
    \includegraphics[width=\textwidth]{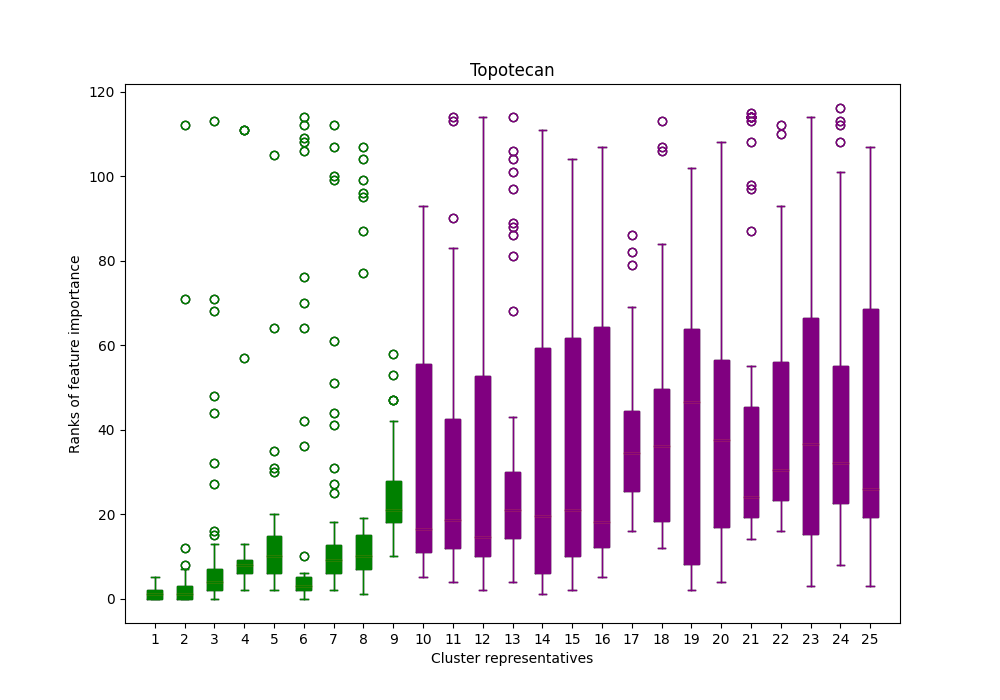}
  \end{minipage}
  \hfill
  \begin{minipage}[b]{0.45\textwidth}
    \includegraphics[width=\textwidth]{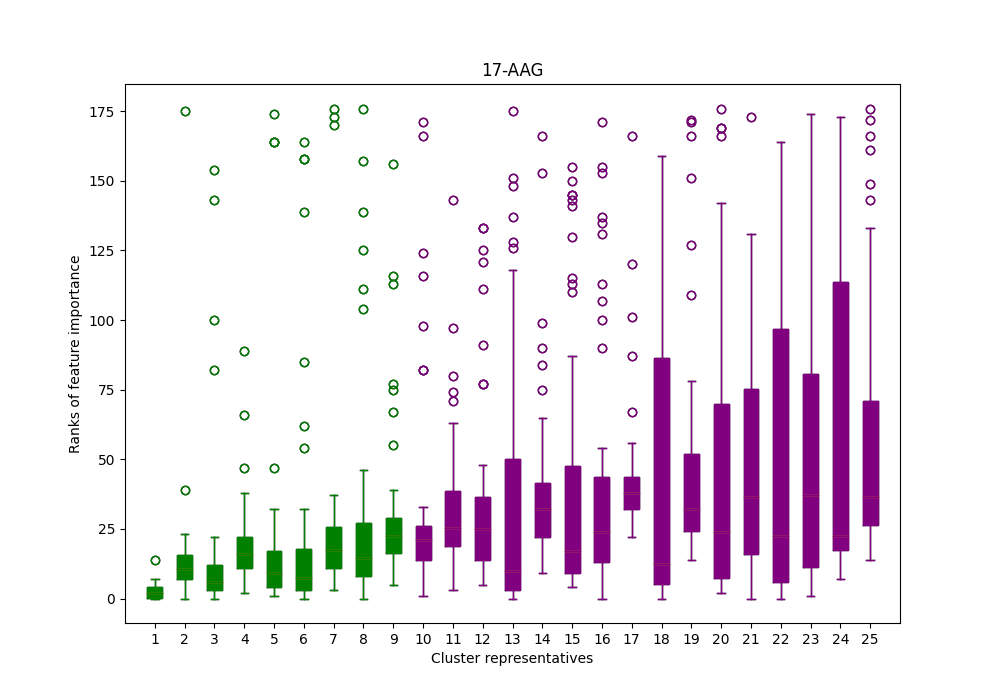}
  \end{minipage}
   \hfill
  \begin{minipage}[b]{0.45\textwidth}
    \includegraphics[width=\textwidth]{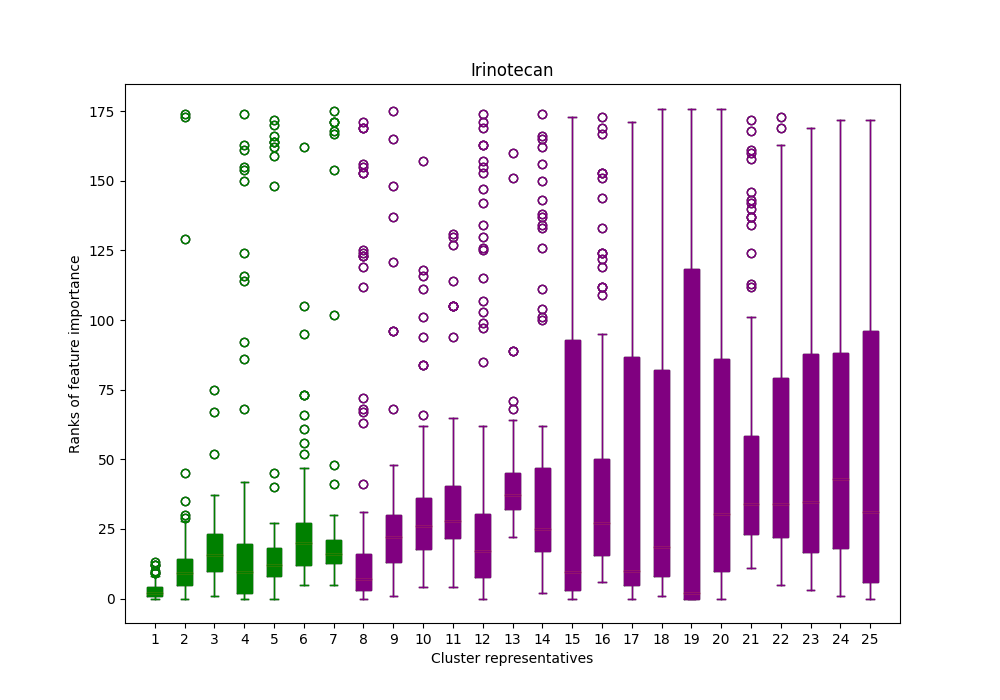}
  \end{minipage}
     \hfill
  \begin{minipage}[b]{0.45\textwidth}
    \includegraphics[width=\textwidth]{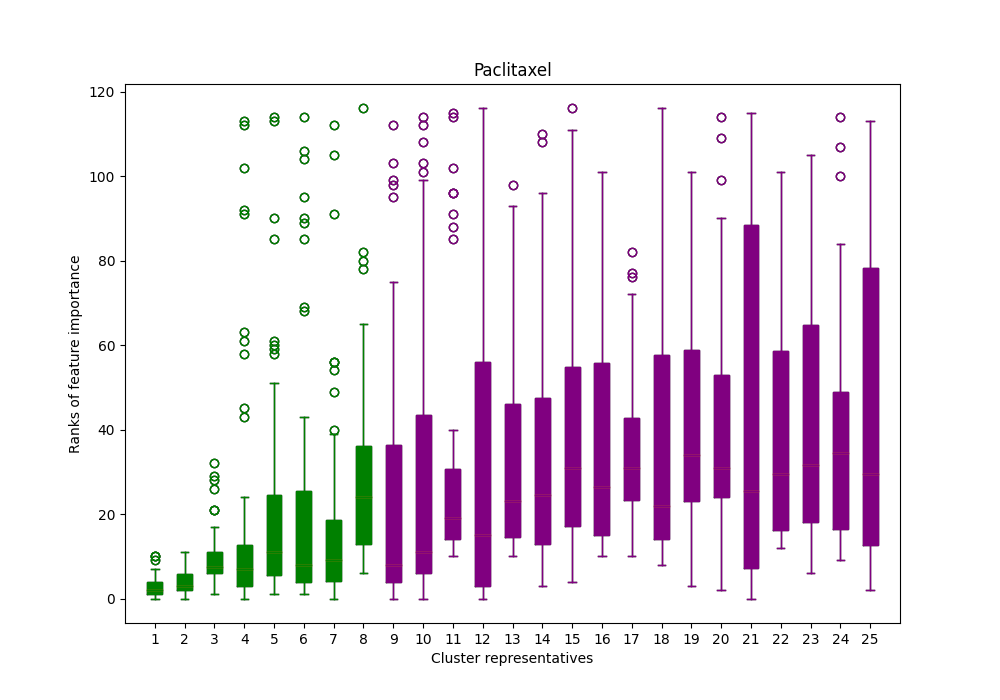}
  \end{minipage}
     \hfill
  \begin{minipage}[b]{0.45\textwidth}
    \includegraphics[width=\textwidth]{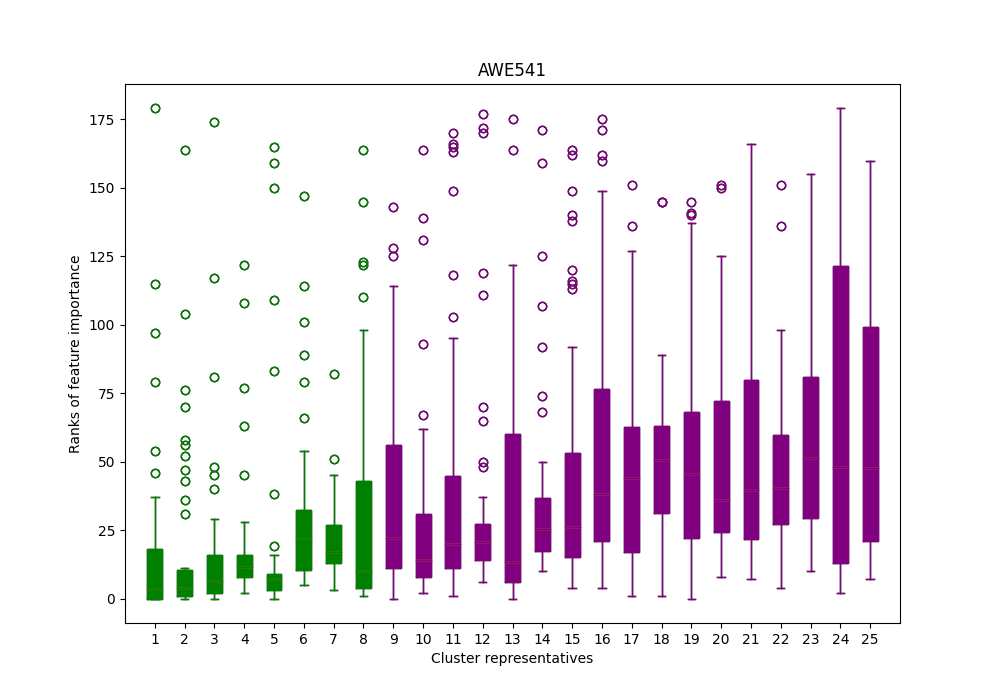}
  \end{minipage}
  \caption{The phase transition property illustrated for the five anticancer drugs considered: (1) Topotecan, (2) 17-AAG, (3) Irinotecan, (4) Paclitaxel, and (5) AWE541, respectively (from top left)}
  \vspace{-.2cm}
  \label{phase_transition_gene}
\end{figure}

The main reason for the phase transition is that, for a null predictor $X_j, j \in S_0^c$, different bootstrap replicates reshuffle its feature importance each time, whereas for a nonnull predictor $X_j, j \in S_0$, the feature importance is much stable in different bootstrap replicates. SciDNet effectively captures this characteristic to identify the null features. As a demonstration, here we present in Figure \ref{phase_transition_gene}, the bootstrap distribution of rank of the importance scores for the top 25 important cluster representatives via box plots. The green and purple colors respectively indicate if the cluster representatives are selected or rejected by the SciDNet. We can observe the phase transition consistently for all five drugs, and SciDNet selects only those important representatives with reduced variability over the bootstrap replicates. 

\section{More Simulation Results} 

\renewcommand{\thetable}{S\arabic{table}}
\begin{table*}
  \caption{Empirical power and observed FDR of SciDNet with standard error in parentheses for gaussian features}
  \centering
  \label{table:simulation2}
  \resizebox{\textwidth}{!}{\begin{tabular}{lcr|cccc|cccc|cccc}
    \toprule
  &  &  & \multicolumn{4}{c|}{Nonlinear Additive }    & \multicolumn{4}{c|}{Nonlinear with interaction} & \multicolumn{4}{c|}{Linear}    \\
    \cmidrule(lr){4-7}\cmidrule(lr){8-11}\cmidrule(lr){12-15}
    $\rho$& $snr$ & $q$ & 0.01 & 0.05 & 0.1 & 0.15 & 0.01 & 0.05 & 0.1 & 0.15 & 0.01 & 0.05 & 0.1 & 0.15\\
    \midrule
    \multirow{6}{*}{$\rho=0.9$} & \multirow{2}{*}{$snr=9:1$} & Power & 0.79 (0.19) & 0.93 (0.11) & 0.96 (0.09) & 0.96 (0.09) & 0.99 (0.06) & 1.00 (0.00) & 1.00 (0.00) & 1.00 (0.00) & 1.00 (0.00) & 1.00 (0.00) & 1.00 (0.00) & 1.00 (0.00)\\
    & & FDR & 0.00 (0.00) & 0.00 (0.02) & 0.01 (0.03) & 0.02 (0.05) & 0.00 (0.00) & 0.02 (0.05) & 0.03 (0.06) & 0.04 (0.08) & 0.00 (0.00) & 0.01 (0.04) & 0.01 (0.05) & 0.01 (0.05)\\
    & \multirow{2}{*}{$snr=8:2$} & Power & 0.59 (0.20) & 0.82 (0.15) & 0.86 (0.14) & 0.87 (0.13) & 0.84 (0.24) & 1.00 (0.03) & 1.00 (0.03) & 1.00 (0.03) & 1.00 (0.00) & 1.00 (0.00) & 1.00 (0.00) & 1.00 (0.00)\\
    & & FDR & 0.00 (0.03) & 0.00 (0.03) & 0.03 (0.07) & 0.04 (0.08) & 0.00 (0.00) & 0.01 (0.03) & 0.02 (0.06) & 0.04 (0.07) & 0.01 (0.05) & 0.02 (0.06) & 0.02 (0.06) & 0.02 (0.06)\\
    & \multirow{2}{*}{$snr=7:3$} & Power & 0.42 (0.20) & 0.65 (0.12) & 0.77 (0.14) & 0.81 (0.14) & 0.72 (0.26) & 0.94 (0.12) & 0.95 (0.12) & 0.96 (0.11) & 1.00 (0.00) & 1.00 (0.00) & 1.00 (0.00) & 1.00 (0.00)\\
    & & FDR & 0.00 (0.00) & 0.00 (0.00) & 0.00 (0.03) & 0.01 (0.05) & 0.01 (0.04) & 0.03 (0.06) & 0.03 (0.07) & 0.05 (0.09) & 0.00 (0.02) & 0.02 (0.05) & 0.02 (0.05) & 0.02 (0.06)\\
    \midrule
    \multirow{6}{*}{$\rho=0.95$} & \multirow{2}{*}{$snr=9:1$} & Power & 0.83 (0.18) & 0.96 (0.08) & 0.98 (0.06) & 0.98 (0.06) & 0.99 (0.04) & 1.00 (0.00) & 1.00 (0.00) & 1.00 (0.00) & 1.00 (0.00) & 1.00 (0.00) & 1.00 (0.00) & 1.00 (0.00)\\
    & & FDR & 0.00 (0.03) & 0.02 (0.05) & 0.03 (0.07) & 0.04 (0.07) & 0.00 (0.00) & 0.04 (0.07) & 0.08 (0.09) & 0.11 (0.10) & 0.00 (0.02) & 0.06 (0.09) & 0.08 (0.10) & 0.11 (0.12)\\
    & \multirow{2}{*}{$snr=8:2$} & Power & 0.60 (0.29) & 0.82 (0.17) & 0.87 (0.14) & 0.89 (0.12) & 0.98 (0.08) & 0.99 (0.05) & 0.99 (0.05) & 0.99 (0.05) & 1.00 (0.00) & 1.00 (0.00) & 1.00 (0.00) & 1.00 (0.00)\\
    & & FDR & 0.00 (0.03) & 0.02 (0.05) & 0.02 (0.06) & 0.03 (0.07) & 0.01 (0.04) & 0.04 (0.07) & 0.08 (0.12) & 0.12 (0.12) & 0.01 (0.03) & 0.05 (0.09) & 0.07 (0.10) & 0.09 (0.11)\\
    & \multirow{2}{*}{$snr=7:3$} & Power & 0.32 (0.22) & 0.61 (0.19) & 0.79 (0.15) & 0.82 (0.15) & 0.80 (0.27) & 0.96 (0.11) & 0.98 (0.05) & 0.98 (0.05) & 0.93 (0.19) & 0.96 (0.11) & 0.97 (0.09) & 0.97 (0.09)\\
    & & FDR & 0.00 (0.00) & 0.01 (0.04) & 0.01 (0.05) & 0.03 (0.08) & 0.00 (0.02) & 0.04 (0.07) & 0.07 (0.09) & 0.12 (0.10) & 0.00 (0.03) & 0.04 (0.08) & 0.06 (0.08) & 0.07 (0.09)\\

    \bottomrule
  \end{tabular}}
\end{table*}

Here we demonstrate finite sample performance of SciDNet under various linear and nonlinear models with varying multicollinearity level under different signal-to-noise-ratio. 
\subsection{Using Gaussian Features} \label{more_simulation_gauss}
For the high dimensional predictors, n i.i.d. copies are first generated from $X \sim N_p(0, \Sigma)$, where $n=600$, $p=5000$ and the covariance matric $\Sigma$ is chosen as a toeplitz matrix with $\Sigma_{ij}=\rho^{|i-j|}$. The value of $\rho$ is varied to explore different correlation strength. We set the set of truely significant variables $S=\{100,200,300,400,500\}$ with $s=5$. The response $y$ is generated from $y=g(x)+\epsilon$. Here we entertain the following three models: 
\begin{enumerate}
    \item \textbf{Linear}: $g(x)= x_S\beta_S$ with $\beta_S$ generated from $N(2,0.1)$ independently and $\beta_{S^c}=0$,
    \item \textbf{Nonlinear additive}: $g(x)=2x_{100}+ 2x_{200}^3 + e^{x_{300}}+6\sin{x_{400}}+2ReLu(x_{500}^3)$, where ReLu(x)=max(x,0)
    \item \textbf{Nonlinear with interaction}: $g(x)=2x_{100}+ 2x_{200}^3 + e^{x_{300}}+6x_{400}x_{500}$
\end{enumerate}
In each cases, the random noise $\epsilon$ is independently generated from $N(0,\sigma^2)$, where the value of $\sigma^2$ is chosen maintaining the signal-to-noise ratio at the desired level. To this end, we define the signal-to-noise ratio as $snr=\frac{var(g(x))}{\sigma^2}$. Here we consider three levels of $snr=9:1, 8:2$ and $7:3$. Table \ref{table:simulation2} shows that SciDNet continues to maintain satisfactory power while successfully controlling the FDR below the threshold $q=0.01, 0.05, 0.1, 0.15$. The average cluster size is observed at 8.3 for $\rho=0.9$ and 13.4 for $\rho=0.95$.

\subsection{Using Non-gaussian Features}
To check SciDNet's performance under non-gaussian setup, n iid copies of high-dimensional feature vector X are generated from multivariate $t_p(5)$ distribution considering same correlation structure as in the previous section \ref{more_simulation_gauss},  with n=600, p=5000. The remaining simulation setting is consistent with the previous section 2.1.  The performance of SciDNet is presented at table \ref{table:non_gaussian} which is quite analogous to the results of gaussian features.

\renewcommand{\thetable}{S\arabic{table}}
\begin{table*} 
  \caption{Empirical power and observed FDR of SciDNet with standard error in parentheses for non-gaussian features}
  \centering
  \label{table:non_gaussian}
  \resizebox{\textwidth}{!}{\begin{tabular}{lcr|cccc|cccc|cccc}
    \toprule
  &  &  & \multicolumn{4}{c|}{Nonlinear Additive }    & \multicolumn{4}{c|}{Nonlinear with interaction} & \multicolumn{4}{c|}{Linear}    \\
    \cmidrule(lr){4-7}\cmidrule(lr){8-11}\cmidrule(lr){12-15}
    $\rho$& $snr$ & $q$ & 0.01 & 0.05 & 0.1 & 0.15 & 0.01 & 0.05 & 0.1 & 0.15 & 0.01 & 0.05 & 0.1 & 0.15\\
    \midrule
    \multirow{6}{*}{$\rho=0.9$} & \multirow{2}{*}{$snr=9:1$} & Power & 0.68 (0.12) & 0.96 (0.15) & 1.00 (0.09) & 1.00 (0.07) & 0.92 (0.05) & 0.95 (0.02) & 1.00 (0.00) & 1.00 (0.00) & 1.00 (0.00) & 1.00 (0.00) & 1.00 (0.00) & 1.00 (0.00)\\
    & & FDR & 0.00 (0.00) & 0.00 (0.01) & 0.04 (0.01) & 0.07 (0.05) & 0.00 (0.00) & 0.01 (0.06) & 0.04 (0.06) & 0.06 (0.07) & 0.00 (0.00) & 0.01 (0.04) & 0.01 (0.05) & 0.01 (0.05)\\
    & \multirow{2}{*}{$snr=8:2$} & Power & 0.56 (0.24) & 0.86 (0.11) & 0.94 (0.14) & 0.95 (0.13) & 0.76 (0.23) & .93 (0.02) & 1.00 (0.04) & 1.00 (0.03) & 1.00 (0.00) & 1.00 (0.00) & 1.00 (0.00) & 1.00 (0.00)\\
    & & FDR & 0.00 (0.01) & 0.00 (0.01) & 0.06 (0.05) & 0.09 (0.07) & 0.00 (0.00) & 0.00 (0.02) & 0.04 (0.03) & 0.07 (0.07) & 0.01 (0.05) & 0.02 (0.06) & 0.02 (0.06) & 0.02 (0.06)\\
    & \multirow{2}{*}{$snr=7:3$} & Power & 0.42 (0.21) & 0.77 (0.13) & 0.91 (0.16) & 0.94 (0.12) & 0.73 (0.26) & 0.94 (0.12) & 0.95 (0.15) & 0.96 (0.14) & 1.00 (0.00) & 1.00 (0.00) & 1.00 (0.00) & 1.00 (0.00)\\
    & & FDR & 0.00 (0.00) & 0.00 (0.01) & 0.02 (0.03) & 0.06 (0.04) & 0.01 (0.05) & 0.04 (0.06) & 0.05 (0.07) & 0.05 (0.09) & 0.00 (0.02) & 0.02 (0.05) & 0.02 (0.05) & 0.02 (0.06)\\
    \midrule
    \multirow{6}{*}{$\rho=0.95$} & \multirow{2}{*}{$snr=9:1$} & Power & 0.81 (0.19) & 0.95 (0.07) & 0.98 (0.06) & 0.98 (0.07) & 0.99 (0.04) & 0.99 (0.03) & 1.00 (0.00) & 1.00 (0.00) & 1.00 (0.00) & 1.00 (0.00) & 1.00 (0.00) & 1.00 (0.00)\\
    & & FDR & 0.00 (0.01) & 0.03 (0.06) & 0.03 (0.04) & 0.05 (0.03) & 0.00 (0.00) & 0.04 (0.07) & 0.08 (0.09) & 0.09 (0.13) & 0.00 (0.01) & 0.03 (0.05) & 0.07 (0.11) & 0.10 (0.14)\\
    & \multirow{2}{*}{$snr=8:2$} & Power & 0.65 (0.29) & 0.84 (0.16) & 0.89 (0.17) & 0.89 (0.12) & 0.94 (0.07) & 0.97 (0.04) & 0.99 (0.07) & 0.99 (0.07) & 1.00 (0.00) & 1.00 (0.00) & 1.00 (0.00) & 1.00 (0.00)\\
    & & FDR & 0.00 (0.03) & 0.01 (0.05) & 0.04 (0.06) & 0.05 (0.06) & 0.01 (0.03) & 0.04 (0.04) & 0.07(0.14) & 0.11 (0.11) & 0.01 (0.02) & 0.05 (0.09) & 0.06 (0.14) & 0.09 (0.11)\\
    & \multirow{2}{*}{$snr=7:3$} & Power & 0.47 (0.22) & 0.64 (0.17) & 0.75 (0.19) & 0.87 (0.11) & 0.82 (0.27) & 0.95 (0.10) & 0.98 (0.04) & 0.98 (0.02) & 0.95 (0.10) & 0.96 (0.11) & 0.97 (0.08) & 0.97 (0.06)\\
    & & FDR & 0.00 (0.00) & 0.02 (0.04) & 0.02 (0.04) & 0.04 (0.09) & 0.00 (0.01) & 0.04 (0.05) & 0.09 (0.03) & 0.13 (0.09) & 0.00 (0.02) & 0.04 (0.05) & 0.05 (0.08) & 0.08 (0.07)\\

    \bottomrule
  \end{tabular}}
\end{table*}

\section{Performance of existing feature selection methods in the presence of high multicollinearity} \label{knockoff_linear}
In this section, we present a numerical illustration of performance of several recently proposed nonlinear FDR-controlled feature selection algorithms. The predictors are first generated from $X_i \sim N_p(0, \Sigma), i=1,2,\dots,n$, for multiple combination of $(n,p)$ and the covariance matric $\Sigma$ is chosen as a toeplitz matrix with $\Sigma_{ij}=\rho^{|i-j|}, \rho=0.1,0.5, \text{ and }0.9$. Under simplistic setting, the response $y$ is generated from $y=x_S\beta_S+\epsilon$, $S=\{5,10,\dots,50\}, |S|=10$, with $\beta_S$ generated from $N(\beta_0,0.1)$ independently and $\beta_{S^c}=0$. The random noise $\epsilon \sim N(0, 1)$. We focus on the Model-X knockoff \citep{Model_X}, SurvNet \citep{survnet}, DeepPINK \citep{deeppink}. For a more rigorous analysis, we consider two different versions of Model-X knockoff  - (1) Model-X-Estimated, where the knockoffs are generated using an estimated multivariate gaussian distribution and (2) Model-X-True,  where the knockoffs are generated using the true data generating multivariate gaussian distribution mentioned above. For the knockoff generation, we consider the equicorrelated construction using the R package \href{https://cran.r-project.org/web/packages/knockoff/index.html}{knockoff: The Knockoff Filter for Controlled Variable Selection}. To implement the SurvNet and DeepPINK, we use the codes mentioned in the respective papers \cite{survnet, deeppink}. We set $q=0.15$ as the FDR control threshold.

Table \ref{table:simulation3} reveals several interesting characteristics. Both Model-X-Estimated and  Model-X-True maintains the power-FDR balance under low correlation setup. However with higher multicollinearity, Model-X-Estimated fails to control the FDR below the specified threshold while the Model-X-True controls the FDR efficiently. This disparity indicates Model-X procedure induces inflation in false discoveries if the knockoffs are not generated properly under 'difficult' situation. As expected, the DL-based algorithms, such as SurvNet and DeepPINK work much better in big-n-small-p and low correlation setup but typically fail in other cases, indicating their reduced effectiveness in ultrahigh dimensional data with small sample size.

\begin{table*}
  \caption{Empirical power and observed FDR of various feature selection algorithms with standard error in parentheses}
  \centering
  \label{table:simulation3}
  \resizebox{0.8\textwidth}{!}{\begin{tabular}{lcr|ccc|ccc}
    \toprule
  &  &  & \multicolumn{3}{c|}{$\beta=2$}    & \multicolumn{3}{c|}{$\beta=4$} \\
    \cmidrule(lr){4-6}\cmidrule(lr){7-9}
    & $\left(n,p\right)$ &  & $\rho=$0.1 & $\rho=$0.5 & $\rho=$0.9 & $\rho=$0.1 & $\rho=$0.5 & $\rho=$0.9 \\
    \midrule
    \multirow{2}{*}{Model-X-Estimated} & \multirow{2}{*}{$\left(400,1000\right)$} & Power & 1.00 (0.00) &  1.00 (0.00) &  1.00 (0.00) &  1.00 (0.00) &  1.00 (0.00) &  1.00 (0.00) \\
    & & FDR & 0.13 (0.17) & 0.12 (0.12) & 0.27 (0.18) & 0.11 (0.19) & 0.20 (0.18) & 0.27 (0.20) \\
    \midrule
    \multirow{2}{*}{Model-X-True} & \multirow{2}{*}{$\left(400,1000\right)$} & Power & 1.00 (0.00) &  1.00 (0.00) &  1.00 (0.00) &  1.00 (0.00) &  1.00 (0.00) &  1.00 (0.00) \\
    & & FDR & 0.08 (0.13) & 0.09 (0.12) & 0.14 (0.17) & 0.12 (0.14) & 0.11 (0.13) & 0.08 (0.12) \\
    \midrule
    \multirow{4}{*}{SurvNet} & \multirow{2}{*}{$\left(400,1000\right)$} & Power & 0.27 (0.20) &  0.32 (0.22) &  0.35 (0.24) &  0.49 (0.24) &  0.52 (0.28) &  0.58 (0.29) \\
    & & FDR & 0.31 (0.36) & 0.53 (0.30) & 0.59 (0.23) & 0.21 (0.21) & 0.53 (0.18) & 0.60 (0.17) \\
    & \multirow{2}{*}{$\left(10000,60\right)$} & Power & 0.99 (0.05) &  1.00 (0.00) &  1.00 (0.00) &  1.00 (0.00) &  1.00 (0.00) &  1.00 (0.00) \\
    & & FDR & 0.20 (0.15) & 0.80 (0.02) & 0.78 (0.07) & 0.14 (0.11) & 0.80 (0.02) & 0.56 (0.32) \\
    \midrule
    \multirow{4}{*}{DeepPINK} & \multirow{2}{*}{$\left(400,1000\right)$} & Power & 0.01 (0.02) &  0.03 (0.04) &  0.00 (0.00) &  0.03 (0.04) &  0.01 (0.03) &  0.02 (0.05) \\
    & & FDR & 0.23 (0.40) & 0.35 (0.42) & 0.33 (0.47) & 0.45 (0.44) & 0.24 (0.41) & 0.24 (0.40) \\
    & \multirow{2}{*}{$\left(10000,60\right)$} & Power & 1.00 (0.00) &  1.00 (0.00) &  1.00 (0.00) &  1.00 (0.00) &  1.00 (0.00) &  1.00 (0.00) \\
    & & FDR & 0.18 (0.04) & 0.29 (0.13) & 0.25 (0.11) & 0.17 (0.01) & 0.24 (0.12) & 0.24 (0.12) \\
     \bottomrule
  \end{tabular}}
\end{table*}
\vspace{-.5cm}

\section{Model implementation details and Sensitivity Analysis}\label{implementation}
In this section, we mention the implementation details of SciDNet that we consider for the simulation study and real data analysis. To select the size of the active set $\hat{S}_n$in the screening step, in consistence with \cite{xue2017robust}, we set  $|\hat{S}_n|=[\frac{2n}{log(n)}]$ by selecting the predictors with the top $|\hat{S}_n|$  Henze–Zirkler test statistic $\tilde{w}_k^*$ , where $[z]$ denotes the integer part of z. In all our simulation scenarios, we set r=0.9, the hyperparameter for intra-cluster correlation bound to further integrate highly correlated conditionally dependent clusters. In cleaning step, for LassoNet 100 dimensional one-hidden-layer feed-forward neural network has been used; more detailed model architecture can be found at appendix in \cite{LassoNet}. For creating the compact neighbourhood in the cleaning step, each time we choose the value of $\kappa$ utilizing the phase transition property mentioned in section 2.2 of main manuscript. The feature selection performance of the SciDNet is demonstrated by calculating the average power and cFDR along with their standard error observed in 50 Monte Carlo replications. Each data set is randomly divided into train, validation
and test with a 70-10-20 split. To asses the prediction performance, the test Mean Square Error (MSE) before and after the variable selection has been shown as part of the simulation study. For the prediction model, a 40-dimensional two-hidden-layer feed-forward neural network with ReLU and linear activation function is considered with Adam as optimizer. For the regression tree, we used the bagging for further stabilization, as mentioned in \cite{breiman1996bagging}. The number of leaves and nodes are chosen by minimising the MSE on validation set.

To access the error bar for the sensitivity analysis, we generate a typical data using the polynomial setup (section 3 in main manuscript, with $\beta=2, \sigma^2=1$) and rerun SciDNet 50 times on the same data and set $q=0.1$ as FDR-control threshold. The mean and standard deviations from these 50 replications are following: \textbf{power} = 0.99 (0.01), \textbf{observed FDR} = 0.03 (0.03), \textbf{test error by LassoNet} = 1.048 (0.101), \textbf{test error by SciDNet+RT} = 0.710 (0.002). From computational complexity, after the screening,the bootstraped LassoNet can be run in parallel loop and we conduct all the experiments in a high-performing computing facility with Intel(R) Xeon(R) Platinum 8260 CPU @ 2.40GHz and 4 Tesla V100S. The codes are available at an \href{https://anonymous.4open.science/r/SciDNet-3CA8}{anonymous repository} (https://anonymous.4open.science/r/SciDNet-3CA8).

\section{Important clusters of gene discovered by SciDNet} \label{imp_genes}
\subsection{For CCLE dataset}
Following Table \ref{CCLE_results} presents all the selected cluster of genes by SciDNet for the five anticancer drugs considered. The genes in a single cluster are mentioned in the "\{\}". Previous research on this gene-expression data has revealed several genes as biologically associated with the corresponding drugs.  SciDNet successfully discovers these genes as the top-most important gene associated with the drugs.In Table \ref{CCLE_results}, the selected genes which are confirmed by previous domain research, are highlighted and corresponding references are mentioned in the column 3. 
\begin{table}
      \caption{Drug-sensitive genes identified by SciDNet and confirming references}
      \label{CCLE_results}
      \centering
      \resizebox{\textwidth}{!}{\begin{tabular}{l|c|c|}
      \toprule
     Drug & Selected clusters of genes  &  Confirming references  \\
      
       \midrule
       Topotecan & \textit{\textbf{\{\underline{SLFN11}\}},\{TUFT1,THRB\},\{CDT1,SF3A2,SNRPA\},\{FTH1P10,FTH1\},\{RPL18\},\{KLF5\}}, & \cite{barretina2012cancer} \\
       & \textit{\{RPL11,RPL5P4,RPS8,RPL5,RPL10A,AL162151.3,RPS9,RPL3\}}, & \cite{li2012feature} \\
       & \textit{\{KIF15,CCNA2,LMNB1,KIF22,AC009133.14\},\{MATN2,HSPB8\}}, & \\
       \midrule
       17-AAG & \textit{\{\textbf{\underline{NQO1}},CTD.2033A16.1\},\{BAX\},\{SLC16A3\},\{PHPT1,SH3BP1\}} & \cite{hadley2014use} \\
     & \textit{\{SPCS3,DCTD\},\{CTD.2008A1.2,SORD\},\{NSMCE4A\},\{CSK\}} &  \cite{barretina2012cancer} \\ 
     \midrule
     Irinotecan & \textit{\textbf{\{\underline{SLFN11}\}},\{KIF15,LMNB1,ARHGAP19\},\{TCEANC2\},\{KIF21B\},\{SQSTM1\},\{HDAC11\}} & \cite{barretina2012cancer} \\
     & \textit{\{KHDRBS1,HNRNPA1P35,HMGB2,HNRNPA1,HNRNPA1L2,} & \cite{li2012feature} \\
     & \textit{HNRNPA1P48,HNRNPA1P7,AC021224.1,HNRNPA1P10,RBMX\}} & \\
     \midrule
     PaclitaXel & \textit{\{PARP1,\textbf{\underline{BCL2L1}}\},\{MMP24\},\{DIMT1\},\{RP11.872D17.4,SSRP1,MTA2\},\{DCUN1D3\}} & \cite{Dorman2016-oo}\\
     & \textit{\{RPL10AP6,RPL10A,EEF2,RPL3\},\{ARHGAP11B,ARHGAP11A,BUB1B,CASC5\},\{HCLS1,LCP1\}} & \cite{lee2016treatment}\\
     \midrule
     AEW541 & \textit{\{TCEAL4,\textbf{\underline{MID2}}\},\{E2F6\},\{AC096772.6\},\{SLC44A1\},\{PGM1\}} & \cite{liang2018bayesian}\\

     & \textit{\{ATP8B2,RNF122\}, \{RP11.1017G21.4\},\{ETNK2\},\{NHS\},\{ATG13\}} & \\
    \bottomrule
    \end{tabular}}
\end{table}

\subsection{For Riboflavin dataset}

The Riboflavin production dataset contains much complicated correlation structure than the CCLE data, see Figure 1 in the main manuscript for a visual illustration. As a result, SciDNet has produced much larger cluster of genes compared to the cluster sizes from the CCLE dataset. For example, the average cluster size for CCLR and Riboflavin dataset is respectively 2.5 and 17.78. Following Table \ref{sel_genes} shows the 9 selected cluster of genes selected by SciDNet while the FDR is controlled at $q=0.15$. Additionally, SciDNet discovered the gene \textit{YCIC\_at} as one of the expressive genes related to riboflavin production which was identified by \cite{buhlmann2014high} as a causal gene in this context.
\begin{table}
  \caption{Selected clusters of genes by SciDNet applied in the riboflavin gene data example}
  \label{sel_genes}
  \centering
   \resizebox{0.8\textwidth}{!}{\begin{tabular}{cl}
    \toprule
    Cluster No.    & Genes selected    \\
    \midrule
  \multirow{ 3}{*}{1} & EPR\_at, IOLD\_at, KAPB\_at, PROJ\_at, RPLQ\_at, UREA\_at, \\ & YCGB\_at, YCGM\_at, YCGN\_at,  YCSN\_at, YCGO\_at, YCGT\_at, \\ &  YDBM\_at, YHXA\_at, YKZC\_at, YOAB\_at, YPJB\_at, YUSX\_at, YVFH\_at \\
  \midrule
   \multirow{ 3}{*}{2}   & COMX\_at, CSPC\_at, HAG\_at, MPR\_at, YBDL\_at, YDBM\_at, \\ & YHCB\_at, YJFB\_at, YHFS\_at, YOAB\_at, YODF\_at, YOAC\_at, \\ & YONU\_at, YOTL\_at, YQKI\_at, YQZH\_at, YTEI\_at, YUSV\_at	\\
   \midrule
   \multirow{ 5}{*}{3}	& HIT\_at, KATX\_at, LICH\_at, NASA\_at, OPUCB\_at,  \\ & PHRG\_i\_at, PHRK\_at, ROCB\_at, ROCR\_at, SACB\_at, SPOIIE\_at, \\ & TMRB\_at, YACN\_at, YBBJ\_at, YBGB\_at, YCBF\_at, YFKJ\_at, \\ & YHCS\_at, YHXA\_at, YJBF\_at, YLBA\_at, YLOU\_at, YPUI\_at, \\ &  YQGY\_i\_at, YUKE\_at, YVYD\_at, YXLJ\_at, YXZF\_at \\
   \midrule
 \multirow{ 5}{*}{4} & APPA\_at, BGLS\_at, ccpB\_at, MMR\_at, SIGY\_at, SOJ\_at, \\ & TREA\_at, YBGB\_at, YDGF\_at, YOPR\_at, YQEB\_at, YVCI\_at,  \\ & YVDR\_at, YWBG\_at, YWDE\_at, YWFM\_at, YXBB\_at, YXIL\_at,  \\ & YXIO\_at, YXIQ\_at, YXJA\_at, YXJN\_at, YXLC\_at, YXLD\_at,  \\ & YXLE\_at, YXLF\_at, YXLG\_at, YXLJ\_at, YXZF\_at, YYBF\_at \\ 
 \midrule
 \multirow{ 2}{*}{5} & LYTD\_at, SQHC\_at, XKDE\_at, YFIG\_at, YFIH\_at, YFII\_at,  \\ & YFNC\_at, YHDV\_at, YIST\_at, YJGA\_at, YTCP\_at, YTMP\_at \\
 \midrule
 \multirow{ 3}{*}{6} &	YCDH\_at, YCDI\_at, YCEA\_at, YCIA\_at, YCIB\_at, \textbf{\underline{YCIC\_at}}, \\ &  YDAR\_at, YHZA\_at, YRPE\_at, YTGA\_at, YTGB\_at, YTGC\_at, \\ & YTGD\_at, YTIA\_at, YVQH\_at\\
 \midrule
 \multirow{ 4}{*}{7} &	OPUBD\_at, PHRE\_at, SIPS\_at, YBFF\_at, YDEM\_at, YNAB\_i\_at, \\ & YNAC\_at, YNEK\_at, YOBF\_at, YOKG\_at, YONX\_at, YOPA\_at, \\ & YOPR\_at, YOTL\_at, YPBB\_at, YQZH\_at, YRDA\_at, YRKK\_at, \\ &
YRKL\_at, YTGB\_at, YUXI\_at, YWCE\_at, YWQK\_at, YYDB\_at, YYDF\_i\_at\\
\midrule
8 & ARGB\_at, ARGC\_at, ARGD\_at, ARGJ\_at, CARA\_at, CARB\_at\\
\midrule
9 & PROJ\_at, RPLF\_at, RPLJ\_at, RPLL\_at, RPSN\_at, RPSP\_at, YLQC\_at\\
\bottomrule
  \end{tabular}}
\end{table}

\end{document}